\documentclass[sigconf]{aamas} 

\usepackage{balance} 

\setcopyright{ifaamas}
\acmConference[AAMAS '24]{}{} {}{}
\copyrightyear{2024}
\acmYear{2024}
\acmDOI{}
\acmPrice{}
\acmISBN{}

\newcommand{\comments}[1]{{\color{blue}\textit{$\#$ #1}}}

\newcommand{\cA}{{\mathcal{A}}}

\newcommand{\cD}{{\mathcal{D}}}
\newcommand{\cS}{{\mathcal{S}}}
\newcommand{\cM}{{\mathcal{M}}}

\newcommand{\cN}{{\mathcal{N}}}

\newcommand{\cO}{{\mathcal{O}}}

\newcommand{\cB}{{\mathcal{B}}}

\newcommand{\bQ}{\textbf{Q}}

\newcommand{\bR}{\textbf{R}}
\newcommand{\bX}{\textbf{X}}

\newcommand{\bV}{\textbf{V}}

\newcommand{\bbR}{\mathbb{R}}
\newcommand{\bbE}{\mathbb{E}}

\newcommand{\ELU}{\textsc{ELU}}

\newcommand{\KL}{\textsc{KL}}

\newcommand{\squishend}{
  \end{list}  }

\DeclareMathOperator*{\argmin}{argmin}

\newif\ifnotes\notestrue
\def\htien#1{}

 \author{The Viet Bui}
 \affiliation{
   \institution{Singapore Management University}
   \city{Singapore}
   \country{Singapore}}
 \email{tvbui@smu.edu.sg}

 \author{Tien Mai}
 \affiliation{
   \institution{Singapore Management University}
 \city{Singapore}
 \country{Singapore}}
 \email{atmai@smu.edu.sg}

\author{Thanh H. Nguyen}
\affiliation{
	\institution{University of Oregon}
	\city{Eugene, Oregon}
	\country{United States}}
\email{thanhhng@cs.uoregon.edu}
\usepackage{graphicx}
\usepackage[linesnumbered,ruled,vlined]{algorithm2e}
\usepackage{subcaption}
\usepackage{multirow}
\usepackage{color, colortbl}
\usepackage{rotating}
\usepackage{setspace}

\newcommand{\showinstance}[4][0.25]{
\begin{minipage}{#1\textwidth}
    \centering
    \includegraphics[width=1.0\textwidth,trim={#2 0 0 0},clip]{graphs/#3.pdf}
    \ifx #4\empty \else \text{\detokenize{#4}} \fi
    \captionsetup{justification=centering}
\end{minipage}
}
\newcommand{\showinstancex}[4][0.25]{
\begin{minipage}{#1\textwidth}
    \centering
    \includegraphics[width=1.0\textwidth,trim={#2 #4 0 0},clip]{graphs/#3.pdf}
    \captionsetup{justification=centering}
\end{minipage}
}
\newcommand{\showlegend}[4][0.25]{
\begin{minipage}{#1\textwidth}
    \centering
    \includegraphics[width=1.0\textwidth,trim={#2cm #3cm 0 0},clip]{graphs/legend_#4.pdf}
\end{minipage}
}

\newcommand{\showsmactable}[2]{
\begin{table*}[ht]
\caption{Results on SMACv2,  the number of expert trajectories is: #1}
\begin{tabular}{cccccccc}
\toprule
\multirow{2}{*}{Methods} & \multicolumn{2}{c}{Protoss} & \multicolumn{2}{c}{Terran} & \multicolumn{2}{c}{Zerg} & \multirow{2}{*}{Average} \\ 
\cmidrule(l{0.4em}r{0.4em}){2-3}
\cmidrule(l{0.4em}r{0.4em}){4-5}
\cmidrule(l{0.4em}r{0.4em}){6-7}
& \multicolumn{1}{c}{\textit{5\_vs\_5}} & \textit{10\_vs\_10} & \multicolumn{1}{c}{\textit{5\_vs\_5}} & \textit{10\_vs\_10} & \multicolumn{1}{c}{\textit{5\_vs\_5}} & \textit{10\_vs\_10} & \\ \midrule
#2
\bottomrule
\end{tabular}
\end{table*}
}

\newcommand{\showminertable}[2]{
\begin{table*}[ht]
\caption{Results on Miner,  the number of expert trajectories is: #1}
\begin{tabular}{ccccc}
\toprule
\multirow{2}{*}{Methods} & \multicolumn{3}{c}{Miner}                                         & \multirow{2}{*}{Average} \\ \cmidrule{2-4}
                     & \multicolumn{1}{c}{\textit{easy}}   & \multicolumn{1}{c}{\textit{medium}} & \textit{hard}   &                          \\ \midrule
#2
\bottomrule
\end{tabular}
\end{table*}
}

\newcommand{\showmpetable}[2]{
\begin{table*}[ht]
\caption{Results on MPEs,  the number of expert trajectories is: #1}
\begin{tabular}{ccccc}
\toprule
\multirow{2}{*}{Methods} & \multicolumn{3}{c}{Simple} & \multirow{2}{*}{Average} \\ \cmidrule{2-4}
& \multicolumn{1}{c}{\textit{reference}} & \multicolumn{1}{c}{\textit{spread}} &\textit{speaker\_listener} & \\ \midrule
#2
\bottomrule
\end{tabular}
\end{table*}
}

\newcommand{\red}[1]{\textcolor{red}{\textbf{#1}}}

\acmSubmissionID{815}

\title[AAMAS-2024 Formatting Instructions]{Inverse Factorized Q-Learning for Cooperative Multi-agent Imitation Learning}

\usepackage{algpseudocode}

\begin{abstract}
This paper concerns imitation learning (IL) (i.e, the problem of learning to mimic expert behaviors from demonstrations) in cooperative multi-agent systems.
The learning problem under consideration poses several challenges, characterized by high-dimensional state and action spaces and intricate inter-agent dependencies. In a single-agent setting, IL has proven to be done efficiently through an inverse \textit{soft-Q} learning process given expert demonstrations. However, extending this framework to a multi-agent context introduces the need to simultaneously learn both local value functions to capture local observations and individual actions, and a joint value function for exploiting centralized learning. 
In this work, we introduce a novel multi-agent IL algorithm designed to address these challenges. Our approach enables the
centralized learning by leveraging mixing networks to aggregate  decentralized Q functions. A main advantage of this approach is that  the weights of the mixing networks can be trained using information derived from global states.  We further establish conditions for the mixing networks under which the multi-agent objective function exhibits convexity within the Q function space.
We present  extensive experiments conducted on some challenging  competitive and cooperative multi-agent game environments,
including an advanced version of the \textit{Star-Craft} multi-agent
challenge (i.e., \textit{SMACv2}), which demonstrates the effectiveness of our proposed algorithm compared to existing state-of-the-art multi-agent IL algorithms.

\end{abstract}


\newcommand{\BibTeX}{\rm B\kern-.05em{\sc i\kern-.025em b}\kern-.08em\TeX}

\begin{document}


\pagestyle{fancy}
\fancyhead{}


\maketitle 


\section{Introduction}

Imitation learning (IL) is a powerful approach for sequential decision making in complex high-dimensional environments in which agents can learn desirable behavior by imitating an expert. There is a wide range of applications of IL in several real-world domains, including healthcare~\cite{wang2020adversarial,shah2022learning} and autonomous driving~\cite{zhou2021exploring,hawke2020urban,le2022survey}. A rich body of literature on IL focuses on simple single-agent settings~\cite{abbeel2004apprenticeship,ho2016generative,fu2017learning,ziebart2008maximum,ross2010efficient,garg2021iq}. Existing approaches on this line of work are, however, not directly suitable for multi-agent settings due the interactive nature of multiple agents, making the environment non-stationary to individual agents. Recently, researchers develop new multi-agent imitation learning methods that are tailored to either cooperative or non-cooperative (or both) settings~\cite{song2018multi,yu2019multi,le2017coordinated,bhattacharyya2018multi,jeon2020scalable,vsovsic2016inverse}.  

Among these existing multi-agent IL works, leading methods~\cite{song2018multi,yu2019multi} explore equilibrium solution concepts for Markov games including Nash~\cite{hu1998multiagent} and logistic stochastic best response~\cite{yu2019multi}. Findings on underlying properties of these solution concepts are then incorporated into extending the single-agent imitation learning models for multi-agent settings. While these multi-agent IL methods shows promising results on some cooperative and competitive multi-agent tasks, they still face difficulties in training in practice, which is the result of the underlying adversarial optimization process inherent from original single-agent IL methods~\cite{ho2016generative,fu2017learning}. Indeed, this adversarial optimization involves biased and high variance gradient estimators, leading to a highly unstable learning process.

Our work studies IL in cooperative multi-agent settings. We leverage recent advanced findings in both single-agent IL and cooperative multi-agent reinforcement learning (MARL) to build a unified multi-agent IL algorithm, named Multi-agent Inverse Factorized Q-learning (MIFQ). 
Essentially, MIFQ is built upon inverse soft Q-learning (IQ-Learn)~\cite{garg2021iq} --- a leading single-agent IL method of which advantage is to learn a single Q-function that implicitly defines both reward and policy functions, thus avoiding adversarial training. To adapt inverse soft Q-learning for multi-agent settings, we then employ value-function factorization with mixing networks to model both the reward and state-value functions involves in the learning objective, inspired by the leading cooperative MARL method, QMIX~\cite{rashid2020monotonic}. That is, individual state-value $V$-function and reward $R$-function of the agents are combined via two different mixing networks into joint $V$-values and $R$-values, which are then incorporated into the learning objective. This novel integration enables training decentralised imitation policies for agents in a centralised fashion, following the well-known paradigm of centralised training with decentralised execution in cooperative MARL~\cite{oliehoek2008optimal,kraemer2016multi}.

An important research question arising from this integration is that whether the objective function in multi-agent inverse soft Q-learning is still convex or not given the complication of mixing networks involved together with the value-function factorization. We remark that this convexity property is important as it guarantees the optimization objective is well-behaved and has an unique optimization solution, leading to the effectiveness of original single-agent IL method, IQ-Learn~\cite{garg2021iq}. As our second contribution, we provide new theoretical results, showing that the objective remains convex in local Q-values of the agents when the mixing networks are convex and non-decreasing in the corresponding local state and reward values. Following with this theoretical finding, we show that using hyper-networks (that take global states as input and then output weights for mixing networks) with the commonly-used general two-layer neural network structures~\cite{rashid2020monotonic} will guarantee the convexity of the learning objective within the Q-function space. 

Finally, we conduct extensive experiments in various multi-agent domains, including: SMACv2~\cite{ellis2022smacv2}, Gold Miner~\cite{Miner}, and MPE (Multi Particle Environments)~\cite{lowe2017multi}. We show that our algorithm MIFQ outperforms other baseline algorithms in all these environments. It is important to note that our experiments with SMACv2 mark the first time imitation learning  algorithms are employed and evaluated on such a challenging multi-agent environment.

\section{Related Work}

\textbf{Single-Agent Imitation Learning. }There is a rich body of existing works focusing on generating policies that mimic an expert's behavior given data of expert demonstrations in single-agent settings. A classic approach is behavioral cloning which casts IL as a supervised learning problem, attempting to maximize the likelihood of the expert's trajectories~\cite{ross2010efficient,pomerleau1991efficient}. This approach, while simple, requires a large amount of data to work well due to its compounding error issue. An alternative approach is to recover the reward function (either implicitly or explicitly) for which the expert's policy is optimal~\cite{fu2017learning,ho2016generative,reddy2019sqil,kostrikov2019imitation,finn2016connection}. Leading methods~\cite{ho2016generative,fu2017learning} follow an adversarial optimization process (which is similar to GAN~\cite{goodfellow2014generative}) to train the imitation policy and reward function alternatively. These adversarial training-based methods, however, suffer instability. A more recent work by~\cite{garg2021iq} overcomes this instability issue by introducing an inverse soft-Q learning process, i.e., learning a single Q-function from which corresponding policy and reward functions can be inferred. For a comprehensive review of literature on this topic, we refer readers to the survey article in~\cite{arora2021survey}.

\textbf{Multi-Agent Imitation Learning.} Most of single-agent IL works, however, do not apply directly to multi-agent settings, mainly due to the non-stationarity of multi-agent environments. Literature on multi-agent IL is rather limited. A few works study multi-agent IL either in cooperative environments~\cite{barrett2017making,le2017coordinated,vsovsic2016inverse,bogert2014multi} or competitive environments~\cite{lin2014multi,reddy2012inverse,song2018multi,yu2019multi}. Recent leading methods employ equilibrium solution concepts in Markov games such as Nash equilibrium to extend some existing IL methods to the multi-agent settings~\cite{song2018multi,yu2019multi}. However, these methods still suffer the instability challenge during training as they still rely on adversarial training. Our work focuses on multi-agent IL in a fully cooperative Dec-POMDP environment. We utilize the idea of inverse soft-Q learning in single-agent IL~\cite{garg2021iq} to avoid adversarial training. We extend this idea to the multi-agent settings under the paradigm of centralized learning decentralized execution from MARL, allowing an efficient and stable learning process. It is worth noting that \cite{bui2023mimicking} also develops an inverse soft Q-learning procedure for enhancing a PPO-based MARL algorithm. Their IL method differs from our algorithm, as it is designed specifically for the situation that  actions are missing in the demonstrations, and only focuses on decentralized learning.

\textbf{Multi-Agent Reinforcement Learning (MARL).} 
The literature  of MARL encompasses a number of advanced methods, including both centralized and decentralized algorithms. While centralized algorithms \citep{claus1998dynamics} focus on learning a unified joint policy that governs the collective actions of all agents, decentralized learning, pioneered by  \citep{littman1994markov}, involves optimizing each agent's local policy independently.
Additionally, there exists a category of algorithms known as \textit{centralized training and decentralized execution} (CTDE). For instance, methods detailed in \cite{lowe2017multi} and \cite{foerster2018counterfactual} employ actor-critic architectures and train a centralized critic that takes global information into account. Value-decomposition (VD) methods \citep{sunehag2017value,rashid2020monotonic}, represent the joint Q-function as a function of agents' local Q-functions. QMIX, as introduced by \citep{rashid2020monotonic}, offers a more advanced VD method for consolidating agents' individual local Q-functions through the utilization of mixing and hyper-network \citep{ha2017hypernetworks} concepts. 
There are also other policy gradient based  MARL algorithms. For instance, \citep{de2020independent} develops independent PPO (IPPO), a decentralized MARL, that can achieve high success rates in several hard SMAC maps. \citep{yu2022surprising} develops MAPPO, a PPO version for MARL that achieves  SOTA results on several tasks,  and recently, \cite{bui2023mimicking} proposes IMAX-PPO, an enhanced PPO algorithm for MARL. Our work utilizes MAPPO to train our expert and generate expert demonstrations. We also employ a hyper-network architecture \citep{ha2017hypernetworks} to facilitate centralized learning, following a similar approach in \citep{rashid2020monotonic}.


\section{Preliminaries}
\subsection{Cooperative Multi-agent Reinforcement Learning (Coop-MARL)}
A multi-agent cooperative system can be described as 
a decentralized partially observable Markov decision process (Dec-POMDP), 
defined by a tuple $\{\cS, \cO, \cN, \cA, P, R \}$~\cite{oliehoek2016concise}, where $\cS$ is the set of global states shared by all the agents, $\cO$ is the set of local observations of agents, and $\cN$ is all the agents. In addition, $\cA  = \prod_{i\in \cN_\cA} \cA_i$ is the set of joint actions of all the agents, $\cA_i$ is the set of actions of an agent $i\in \cN$, and $P$ is the transition dynamics of the multi-agent environment. Finally, in Coop-MARL, all agents share the same reward function, $R$, that can take inputs as global states and actions of all the agents and return the corresponding rewards. 

At each time step where the global state is $S$, each ally agent $i\in \cN$ takes an action $a_{i} \in\cA_i$ according to a policy $\pi_i(a_{i}\mid o_{i})$, where $o_{i}$ is the local {observation} of agent $i$. The joint action is  defined as
$A = \{a_{i}\mid i\in \cN\}$ and the joint policy is defined accordingly:
\[
\Pi(A\mid S) = \prod\nolimits_{i\in \cN} \pi_i(a_i\mid o_i). 
\]
After all the agent actions are executed, the global state is updated to a new state $S' \in \cS$ with the transition probability $P(S'\mid A,S)$. The objective  of the Coop-MARL problem is to find a joint policy 
 $\Pi(\cdot\mid S)= \prod_i\pi_i(\cdot\mid o_i)$  that maximizes the expected long-term joint reward, formulated as follows:
 \[
  \max\limits_{\Pi}\bbE_{\Pi}\Big[\sum\limits_{t=0}^\infty  R(A_t,S_t)\Big]
 \]
\subsection{Imitation Learning}
In imitation learning (IL), the objective is to recover an expert reward or expert policy function from some expert demonstration data. Several IL methods were developed for single-agent settings~\cite{abbeel2004apprenticeship,ho2016generative,fu2017learning,ziebart2008maximum,ross2010efficient,garg2021iq}. In general, we can directly extend these methods to a fully-observable cooperative multi-agent setting by considering global states and joint actions of the agents in a similar fashion as in the single-agent settings. Let's denote a set of expert demonstrations as $\cD^E = \{\tau = \{(A_t,S_t),~ t = 0, 1...\}\}$ where $A_t$ is the joint action of all agents and $S_t$ is the global state at time step $t$. In the following, we describe some popular IL approaches accordingly. 

\subsubsection{Behavioral Cloning} A classical approach for imitation learning is Behavioral Cloning (BC),  where the objective is to maximize the likelihood of the demonstrations: 
\[
  \max_{\Pi} \sum_{\tau \in \cD^E}\sum_t \ln \big(\Pi(A_t\mid S_t)\big)
\]
It is commonly known that BC has a strong theoretical foundation, as it can guarantee
to return the exact expert policy. However, BC ignores the environment's  dynamics and only works well with offline learning. As a result, it often
requires a huge number of samples to achieve a desired performance~\cite{ross2011reduction}. 

\subsubsection{Distribution Matching}State-of-the-art (SOTA) IL algorithms are typically based on distribution matching.  Specifically, let $\rho^{\Pi}(S,A)$ be the occupancy measure of visiting the state $S$ and joint action $A$, under the joint policy $\Pi$, defined as follows:
\begin{align*}
    & \rho^{\Pi}(S,A) = \Pi(S\mid A) \prod_{t=0}^{\infty} \gamma^t P(S_t= S\mid \Pi)
\end{align*}

The objective is to minimize a divergence between the occupancy measure of the learning policy $\rho^\Pi$ and that of the expert's policy $\rho^{\Pi^E}$. For instance, the imitation learning can be done by solving: 
\[
\min_{\Pi} \Big\{\KL\left(\rho^\Pi \Big\| \rho^{\Pi^E}\right)\Big\} =\min_{\Pi}\bbE_{(S,A)\sim \rho^{\Pi}}\Big[\ln \frac{\rho^{\Pi^E}(S,A)}{\rho^{\Pi}(S,A)}\Big]
\]
IL algorithms based on distributional matching include some
SOTA IL methods such as adversarial IL 
\citep{ho2016generative,fu2017learning}
or IQ-Learn \citep{garg2021iq}.



\section{Multi-agent Inverse  Q-Learning}
As mentioned previously, our new algorithm is built upon an integration of recent advanced findings in imitation learning (i.e., the new leading IL algorithm IQ-Learn~\cite{garg2021iq}) and in multi-agent reinforcement learning (i.e., the SOTA algorithm QMIX~\cite{rashid2020monotonic}). In the following, we first present the main idea of IQ-Learn, with some direct centralized and decentralized adaptations to multi-agent settings. We then discuss key shortcomings of such simple adaptations in the context of a Dec-POMDP environment. Finally, we present our new algorithm which tackles all those shortcomings. 

\subsection{Inverse Soft Q-Learning}
\label{ssec.iqlearn.basics}
\subsubsection{Centralized Inverse Q-Learning}In a fully-observable cooperative multi-agent setting, single-agent IL algorithms including IQ-Learn can be directly applied by using global states and joint actions of the agents in a similar fashion as in single-agent settings. Given demonstration samples $\cD^E = \{\tau = \{(A_t,S_t),~ t = 0, 1...\}\}$ where $A_t$ is the joint action and $S_t$ is the global state at  step $t$, the goal of IQ-Learn is to solve the following maximin problem, which is also the objective of adversarial learning-based IL approaches~\cite{ho2016generative,fu2017learning}: 
\begin{align}
\max\limits_{R} \min\limits_{\Pi} \Big\{L(R,\Pi) &= \bbE_{(S,A)\sim \rho^{E}} \big[R(S,A)\big] \nonumber \\
&- \bbE_{\rho^\Pi}\big[R(S,A)\big] -\bbE_{\rho^\Pi}\big[\ln \Pi(S,A)\big] \Big\} \label{prob:IQ-maximin}   
\end{align}
where $\rho^E$ is the  occupancy measure given by the expert policy $\pi^E$  and $\bbE_{\rho^\Pi}\big[\ln \Pi(S,A)\big]$ is the entropy regularizer. A typical approach to solve this maximin problem is to run an adversarial optimization process over rewards and policies, of which idea is similar to generative adversarial networks~\cite{ho2016generative,fu2017learning}. However, such an approach suffers instability, a well-known challenge of adversarial training. 

In a more recent work~\citep{garg2021iq}, Garg et al. introduce a new IL algorithm, IQ-Learn, which avoids adversarial training by learning a single soft Q-function, of which definition is provided below.
\begin{definition}[Soft Q-function~\cite{garg2021iq}]
    For a reward function R and a policy $\Pi$, the soft Bellman operator $\cB^{\Pi}$ is defined as: $$(\cB^{\Pi}Q)(S,A) = R(S, A) + \gamma \mathbb{E}_{S'\sim P(\cdot\mid S,A)}V^{\Pi}(S') $$ where $V^{\Pi}(S) = \mathbb{E}_{A\sim \Pi(\cdot\mid S)} \big[Q(S, A)-\log \Pi(A\mid S)\big]$. The soft Bellman operator is contractive and defines a unique soft Q-function for the reward function $R$, given as $Q = \cB^{\Pi}Q$.
\end{definition}

Essentially, Garg et al. show that the above maximin problem is equivalent to the following single minimization problem which only requires optimizing over the soft Q-function:\footnote{Here we convert the original maximization formulation in \cite{garg2021iq} into minimization for the sake of our later analyses on the impact of the mixing networks on the convexity of the objective function within the local Q-function space.}  
\begin{align}
    \min\nolimits_{Q}\Big\{J(Q) &=  \bbE_{(S,A)\sim\rho^E} \Big[  \gamma \bbE_{S' \sim P(\cdot\mid S,A)} \big[V^Q(S')\big]- Q(S,A) \Big]\nonumber\\
   &\qquad + (1-\gamma)\bbE_{S^0} \big[V^Q(S^0)\big]\Big\}\label{prob:IQ-global}
\end{align}
where $Q:\cS\times \cA \rightarrow \bbR$ is the soft Q-function and $S^0$ is the initial state. Importantly, the loss functions $J(Q)$ is shown to be convex in $Q$, making the IQ-Learn advantageous to use~\citep{garg2021iq}.

Given $Q$, the optimal policy can be computed accordingly~\citep{garg2021iq}:
\begin{align*}
    & \Pi^Q(A\mid S) = \frac{1}{Z_S} \exp(Q(S,A))
\end{align*}
where $Z_S$ is the normalization factor. 
Finally, the state-value $V^Q(S)$, $S\in \cS$ can be computed as follows: 
\[
V^Q(S) = V^{\Pi^Q}(S) = \ln\Big(\sum\nolimits_{A}\exp(Q(S,A))\Big)
\]

A shortcoming of the above centralized IQ-Learn approach is that it not scalable in complex multi-agent settings since the joint action space grows exponentially in the number of agents. Furthermore, it requires full observations for agents, which is not applicable in a Dec-POMDP environment with only local partial observations. 
\subsubsection{Independent Inverse Q-Learning}\label{sec:IndependentIQ}
An alternative approach to overcome the shortcomings of centralized IQ-Learn is to consider a separate IL problem for each individual agent, taking into account local observations of that agent. Specifically, one can set the objective to recover   a local Q function $Q_i(a_i\mid o_i)$, as a function of a local observation $o_i$ and local action $a_i$,  for each agent $i\in \cN$. The local IQ-Learn loss function can be formulated as follows: 
    \begin{align}
    \min\nolimits_{Q_i}\Big\{J_i(Q_i) &=  \bbE_{(o_i,a_i)\sim \cD^E} \Big[   \gamma \bbE_{o'_i \sim P(\cdot\mid S,A)} \big[V^Q_i(o'_i)\big]  - Q_i(o_i,a_i)\Big]\nonumber\\
   &\qquad + (1-\gamma)\bbE_{o^0_i} \big[V^Q_i(o^0_i)\big]\Big\} \label{prob:IQ-local}
\end{align}
where $V^Q_i(o_i) = \ln\Big(\sum_{a_i\in \cA_i}\exp\big(Q(o_i,a_i)\big)\Big)$. Here, $o'_i \sim P(\cdot\mid S,A)$ is equivalent to $o'_i$ is the local observation of agent $i$ corresponding to the new state $S'\sim P(\cdot\mid S,A)$.

Let $Q^*_i(o_i,a_i)$ be an output of the training $\min_{Q_i} J_i(Q_i)$, then a local reward and policy function can be recovered~\citep{garg2021iq} as follows:  
\begin{align}
    r^*(o_i,a_i) &=  Q^*_i(o_i,a_i) - \gamma \bbE_{o'_i \sim P(\cdot|S,A)} [V_i^{Q^*}(o'_i)] \label{eq:reward-i}\\
    \pi^*_i(a_i\mid o_i) &=\frac{\exp(Q^*_i(o_i,a_i))}{\sum_{a'_i\in \cA_i}\exp(Q^*(o_i,a'_i))}\nonumber
\end{align}
Let us denote $r^Q_i(o_i,a_i) =  Q_i(o_i,a_i) - \gamma \bbE_{o'_i \sim P(\cdot\mid S,A)} \big[V^Q_i(o'_i)\big]$, then the objective in \eqref{prob:IQ-local} can be  rewritten as follows: 
   \begin{align}
    \min\nolimits_{Q_i}\!\Big\{J_i(Q_i) &=  \bbE_{(o_i,a_i)\sim \cD^E} \big[\!-r^Q_i(o_i,a_i) \big] + (1-\gamma)\bbE_{o^0_i} \big[V^Q_i(o^0_i)\big] \Big\} \label{prob:IQ-local-r} 
\end{align}
Thus, independent IQ-Learn can be understood as the process of reconstructing a local reward function, denoted as $r^Q_i$ for each agent $i$, with the objective of minimizing the negation of the expected reward generated by the expert's policy. We will leverage this insight to build mixing networks that facilitate centralized learning. 
 
Overall, this approach enables decentralized policies for agents, allowing it to work in a Dec-POMDP environment. However, it faces the challenge of instability due to the non-stationarity of the multi-agent environment. In addition, it does not leverage additional information on global states available during the training process.

\subsection{Inverse Factorized Soft Q-Learning with Mixing Networks}
\begin{figure*}[t!]
    \centering
    \includegraphics[width=1.0\textwidth]{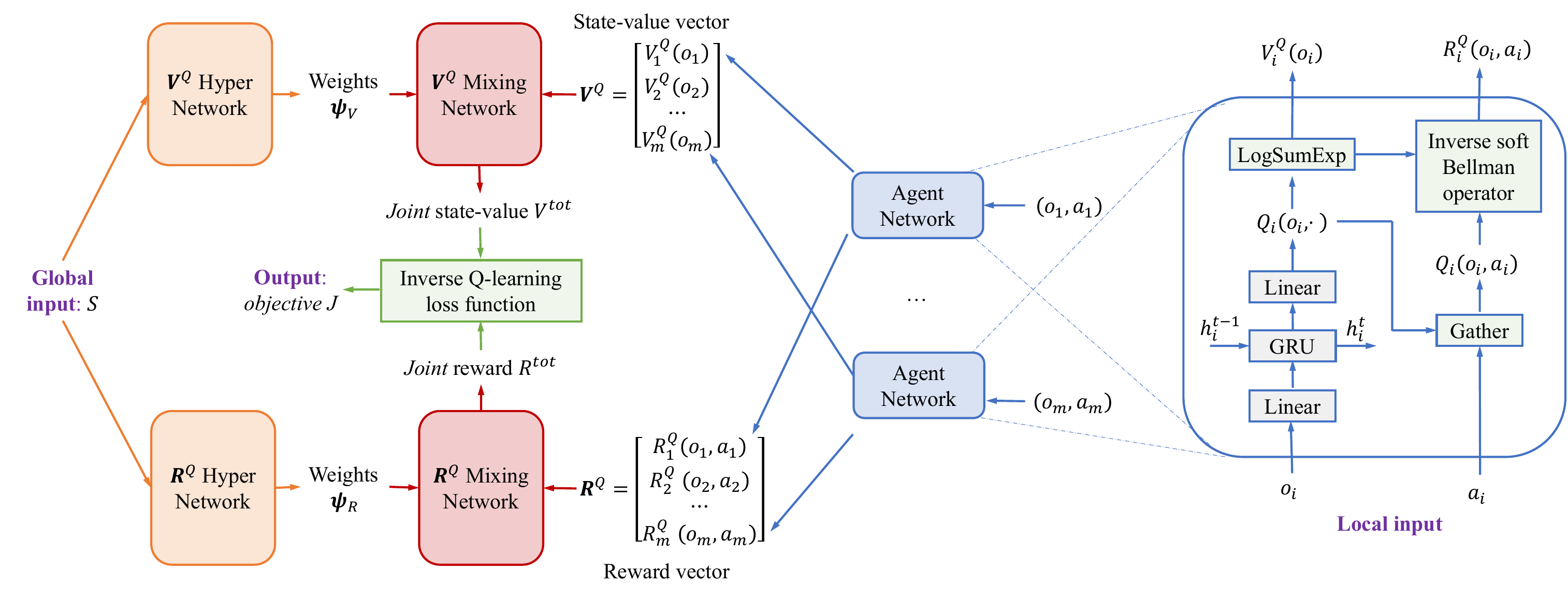}
    \caption{An overview of our multi-agent imitation learning architecture.}
    \label{fig.diagram}
\end{figure*}
We now present our new multi-agent IL approach to learn decentralized value function $Q_i(o_i,a_i)$ in a centralized way.
An overview of our approach is illustrated in Figure~\ref{fig.diagram}. Our key idea involves creating (i) agent local Q-value networks that output local Q-values $Q(o_i, a_i)$ given local information $(o_i,a_i)$ of each agent $i$; and (ii) mixing networks that utilize global information such as global states to combine local values of agents into joint values used to compute the objective of the inverse soft Q-learning. 
In addition, we employ hyper-networks that provide a rich representation for the weights of the mixing networks, allowing us to govern their value ranges. 

\subsubsection{Multi-agent IL Network Architecture}
Overall, our network architecture comprises of three different types of networks:
\paragraph{\normalfont\textbf{Agent local $Q$-value networks.}}To start,  let us define  $\bQ(S,A)$ as a vector $\bQ(S,A)= \big[Q_{1} (o_1, a_1), Q_{2} (o_2, a_2),\cdots, Q_{m} (o_{m}, a_{m})\big]$ of local soft Q-values of agents where $m = |\cN|$ is the number of agents and $(o_i, a_i) \in (S, A),~ i\in \cN$. For an abuse of notations, we also denote by $Q_i(o_i, a_i; \theta_i)$ as the local $Q$-value network of the agent $i$ of which learnable parameter is $\theta_i$.\footnote{In implementation, we consider a shared local $Q$-value network for all agents. } Each local Q-value network takes an input as a pair $(o_i, a_i)$ of a local observation and an action of agent $i$ and then output the corresponding local soft $Q$-value $Q_i(o_i, a_i)=Q_i(o_i, a_i; \theta_i)$. 

We then use this vector $\bQ(S,A)$ to compute the corresponding state-value vector $\bV^Q(S)=\big[V^Q_1(o_1), V^Q_2(o_2),\cdots, V^Q_m(o_m)\big]$ as described in Section ~\ref{ssec.iqlearn.basics}, formulated as follows:
\begin{align*}
    & V^Q_i(o_i) = \ln\Big(\sum\nolimits_{a_i}\exp\big(Q_{i} (o_i, a_i)\big)\Big)
\end{align*}
In addition, let $\bR^Q(S, A)= \big[r^Q_{1} (o_1, a_1), r^Q_{2} (o_2, a_2),\cdots, r^Q_{m} (o_{m}, a_{m})\big]$ as a reward vector comprising of local rewards $r^Q_{i} (o_i, a_i)$ of all agents $i$ where $i \in \cN$. This reward vector can be computed using the inverse soft Bellman operator as follows: 
$$  \bR^Q (S,A) =  \bQ(S,A)-\gamma\bbE_{S'\sim P(\cdot\mid S,A)}\big[\bV^Q(S)\big]$$
The values of $\bV^Q$ and $\bR^Q$ will be passed to the corresponding mixing networks to induce the joint state and reward values $V^{tot}$ and $R^{tot}$ respectively. These two joint values will be then incorporated into computing the objective of the inverse soft Q-learning. 
\paragraph{\normalfont\textbf{State-value and reward mixing networks.}} We create two different mixing networks to combine local state values $\bV^Q$ and local reward values $\bR^Q$ into joint values $V^{tot}$ and $R^{tot}$, respectively. Let's denote these mixing networks as $\cM_{\psi_V}(\cdot)$ and  $\cM_{\psi_R}(\cdot)$. Here $\psi_V$ and $\psi_R$ are corresponding weights (or parameters) of these two mixing networks. In particular, we have: 
\begin{align*}
    &V^{tot}(S) = \cM_{\psi_V}\big(\bV^Q(S)\big)\\
    &R^{tot}(S,A) = \cM_{\psi_R}\big(-\bR^Q(S,A)\big)
\end{align*}

We can now formulate the objective function of our  \textit{multi-agent inverse factorized Q-learning} with respect to the local $Q$-values and these mixing networks, as follows:
\begin{align}
\min_{\bQ} \Big\{ &J(\bQ,\psi_R,\psi_V) = \!\!\!\!\!\!\sum_{(S,A)\in \cD^E}\!\! \!\!\!\!\!R^{tot}(S,A) + (1-\gamma)\bbE_{S^0}\big[V^{tot}(S_0)\big] \Big\}  \label{eq:MICQ-loss-func}
\end{align}
\paragraph{\normalfont\textbf{Hyper-networks.}} Finally, we create two hyper-networks corresponding the two mixing networks. These hyper-networks take the global state $S$ as an input and generate the weights $\psi_V$ and $\psi_R$ of the mixing networks accordingly. The creation of such hyper-networks allows us to have a rich representation of the weights that can be governed to ensure the convexity of the objective $J(\bQ,\psi_R,\psi_V)$ in the local $Q$-values $\bQ(S, A)$ (as we show next). We can write $\psi_V = \psi_V(S; \omega_V)$ and $\psi_R = \psi_R(S; \omega_R)$ where $\omega_V$ and $\omega_R$ denote trainable parameters of the state-value and reward hyper-networks. 

In the end, we can alternatively write the objective function $J(\bQ,\psi_R,\psi_V)$ as $J(\theta, \omega_R, \omega_V)$ when the local soft $Q$-values $\bQ$ are parameterized by $\theta = \{\theta_1,\theta_2,\cdots,\theta_m\}$ and the weights of the mixing networks $(\psi_V,\psi_R)$ are parameterized by $\omega_V$ and $\omega_R$ respectively. We note that we will use either $J(\bQ,\psi_R,\psi_V)$ or $J(\theta, \omega_R, \omega_V)$ depending on the context of our analysis.

\subsubsection{Convexity of Multi-Agent IL Objective Function}
One of key properties that make original IQ-Learn advantageous in single-agent settings is that its loss function is convex in $Q$. Therefore, we aim at exploring conditions under which the loss function in \eqref{eq:MICQ-loss-func} is convex in $\bQ$ in our case.  
For the sake of representation, we will use a common notation $\bX$ which represents either $\bV^Q(S)$ or $-\bR^Q(S, A)$, depending on whether we are referring to the state-value mixing network $\cM_{\psi_V}(\cdot)$ or reward mixing network  $\cM_{\psi_R}(\cdot)$, respectively.\footnote{All of our proofs are in the appendix.}
 \begin{theorem}[Convexity]
 \label{prop:comvex-general}
 Suppose $\cM_{\psi_V}(\bX)$ and $\cM_{\psi_Q}(\bX)$ are convex in $\bX$, non-decreasing in each element $X_i$, then  the objective function $J(\bQ, \psi_R,\psi_V)$ is convex in $\bQ$ .
\end{theorem}
The results in Theorem \ref{prop:comvex-general} are general, in the sense that, as shown below, any feed-forward mixing networks with non-negative weights and nonlinear convex activation functions will satisfy the assumption in Theorem \eqref{prop:comvex-general}, implying the convexity of $J(\bQ,\cdot,\cdot)$.
\begin{theorem}\label{prop:convex-2}
     Any feed-forward mixing networks $\cM_{\psi_R}(\bX)$ and  $\cM_{\psi_V}(\bX)$   constructed with  non-negative weights and nonlinear non-decreasing convex activation functions (such as \textit{ReLU } or \textit{ELU or \textit{Maxout}}) are convex  and non-decreasing in $\bX$.  
\end{theorem}
As a result, Corollary \ref{prop:convex-3} below states that a commonly used two-layers feed-forward network with non-negative weights and convex activation functions will satisfy the conditions in Theorem~\ref{prop:comvex-general} and help preserve the convexity of $J(
\bQ)$.

\begin{corollary}[Convexity under two-layer feed-forward mixing networks]\label{prop:convex-3}
    Suppose the mixing networks $\cM_{\psi_R}(\bX), \cM_{\psi_V}(\bX)$  are two-layer feed-forward neural networks of the form   $
     \sigma(\bX \times |W_1|+ b_1 ) \times |W_2| + b_2 $ where $W_1,W_2,b_1,b_2$ are weight and bias vectors of appropriate sizes and $\sigma$ are some nonlinear convex and non-increasing activation functions (such as \textit{ReLU} or \textit{eLU} or \textit{Maxout}), then $J(\bQ, \psi_R,\psi_V)$ is convex in $\bQ$.
\end{corollary}
The network structure  mentioned in Corollary \ref{prop:convex-3} consists of two layers with convex  activation functions. It also requires that  the  weights of every layer are non-negative. 
Such two-layer mixing structure is widely used in prior  MARL  works~\cite{rashid2020monotonic}. Furthermore, as shown in Theorem~\ref{prop:comvex-general}, the convexity of $J(\bQ,\psi_R,\psi_V)$ still holds under multi-layer feed-forward mixing networks, as long as the activation functions are convex and the weights of each layer are non-negative.  We note that the two-layer structure in Corollary~\ref{prop:convex-3} is sufficient for the mixing network to approximate any monotonic function arbitrarily closely in the limit of infinite width~\cite{dugas2009incorporating}.

\subsubsection{Relation with Independent Inverse Q-Learning. }In its simplest form, mixing networks can be expressed as merely a linear combination of local networks  $Q_i$. We show that in this special case, our inverse factorized soft Q-Learning is equivalent to independent inverse Q-learning. In particular, if the mixing networks are constructed as weighted sums of  $Q_i$ with non-negative weights, it can be demonstrated that minimizing $J(\bQ,\psi_R,\psi_V)$ is essentially equivalent to minimizing each individual local $J_i(Q_i)$ (Proposition \ref{prop:linear-mixing}). One advantage of this linear combination is that it ensures the decentralized objective functions fully align with their centralized counterparts. However, a drawback is the disregard for centralized learning and inter-agent dependencies. Despite this drawback, it underscores the important connection between decentralized and centralized learning. To achieve this, it suffices to have mixing networks $\cM_{\psi_R}(\cdot) $ and $\cM_{\psi_V}(\cdot)$  monotonically increasing. 
With this condition met, for two reward functions $\bR^Q = (r^Q_i,~i \in \cN)$ and $\bR^{Q'} = (r^{Q'}_i,~i \in \cN)$ such that $r^Q_i \geq r^{Q'}_i$ for any $i\in \cN$, we then can see that
$
\cM_{\psi_R}\big(-\bR^Q(S,A)\big) \leq \cM_{\psi_R}\big(-\bR^{Q'}(S,A)\big)
$. 
As a result, the monotonicity ensures that an increase in $r^Q_i$ will result in both a decrease in the local objective $J^i(Q_i)$ and the global objective $J(\bQ,\psi_R,\psi_V)$.  Note that,  to achieve monotonicity, it suffices to ensure that all the weights of the mixing networks non-negative. 

\begin{proposition}[Linear combination]
\label{prop:linear-mixing}
If $\cM_{\psi_R}(\bX)$  and $\cM_{\psi_V}(\bX)$ are weighted sums of $X_i$ with non-negative weighting parameters, i.e., $\cM_{\psi_R}(\bX) = \cM_{\psi_V}(\bX) = \sum_{i}\alpha_i X_i$, $\alpha_i\geq 0$ ($\forall i\in \cN$),  then  $J(\bQ,\psi_R,\psi_V)$ is convex in $\bQ$. Moreover, minimizing $J(\bQ,\psi_R,\psi_V)$ over the $\bQ$ space is equivalent to minimizing  each local loss function $J_i(Q_i)$. That is,
\begin{align*}
    \argmin\limits_\bQ \big\{J(\bQ,\psi_R,\psi_V)\big\} = \big[\argmin\limits_{Q_1} J_1(Q_1),\cdots, \argmin\limits_{Q_m} J_m(Q_m)\big]
\end{align*}

\end{proposition}


\subsection{Practical MIFQ Algorithm}
We now present our practical multi-agent IL algorithm, MIFQ, together with details of our implemented network architecture.
\begin{algorithm}[t!]
\caption{\textbf{M}ulti-agent \textbf{I}nverse \textbf{F}actorized \textbf{Q}-Learning}
\label{algo:mifq}
\textbf{Input:} Environment $ENV$,   parameters $\theta = (\theta_1,\ldots,\theta_m)$ and $(\omega_R,\omega_V)$,
expert's demonstrations $\cD^{E}$, policy replay buffer $D^{replay} = \emptyset$, and learning rates $\lambda_\theta$ and $\lambda_\omega$.\\
\While{a certain number of loops}{
    $ENV.reset()$;\\
     \comments{Collect samples for replay buffer}\\
     \For{\textit{a certain number of collecting steps}}{
         Execute policies $\pi_i \!=\! \text{softmax}~~ Q_{i} (o_i, a_i\!\mid\! \theta_i),~\forall i\!\in\! \cN$ in the environment $ENV$ to collect new transition samples $\{(S, A, S')\}$ and add them to  $\cD^{replay}$;
     }
     
     \comments{Run gradient descent with mini-batches $\sim\cD^{replay}$}\\
    \For{\textit{a certain number of training steps}}{
        Set $\theta \gets \theta - \lambda_\theta \nabla_{\theta} J (\theta, \omega_R,\omega_V)$;\\
        Set $(\omega_R,\omega_V) \gets (\omega_R,\omega_V) \!-\! \lambda_\omega \!\nabla_{\omega_R,\omega_V} J (\theta, \omega_R,\omega_V)$;\\
    }
}
\comments{Recover agent policies and rewards}
\[
\begin{aligned}
\text{Local policies: }\pi_i(a_i|o_i) &= \frac{\exp(Q_i(o_i,a_i,\theta_i))}{\sum_{a'_i\in\cA_i} \exp(Q_i(o_i,a'_i,\theta_i))}, ~~\forall i \in \cN; \\
\text{Local rewards: }r_i(o_i,a_i) &=  Q_i(o_i,a_i) - \gamma \bbE_{o'_i \sim P(\cdot|S,A)} [V_i^{Q}(o'_i)];
\end{aligned}
\]

\end{algorithm}
\subsubsection{Mixing and Hyper Networks}
\label{ssec.mixingnetwork}
We employ the following two-layer feed-forward network structure for our two main mixing networks $\cM_{\psi_R}(\bX)$  and $\cM_{\psi_V}(\bX)$:
\begin{align}
    \cM_{\psi_R}(\bX) &= \ELU(\bX \times |W^R_1|+b^R_1) \times |W^R_2| + b^R_2 \label{eq:M-R} \\
      \cM_{\psi_V}(\bX) &= \ELU(\bX \times |W^V_1|+b^V_1) \times |W^V_2| + b^V_2 \label{eq:M-V}
\end{align}
where $\psi_R = \{W^R_1,W^R_2,b^R_1, b^R_2,\}$ and $\psi_V = \{W^V_1,W^V_2,b^V_1, b^V_2\}$ are the weight and bias  vectors of the mixing networks. The absolute operations $|\cdot|$ are employed to ensure that all the weights are non-negative and, ensuring the convexity of the loss function and the alignment between centralized and local Q functions. Moreover, $\psi_R$ and $\psi_V$ are an output of a \textit{hyper-network} taking the global state $S$ as input. Each hyper-network consists of two fully-connected layers with  a ReLU activation. 
 The output of each hyper-network is a vector, which is then reshaped into a matrix of appropriate size to be used in the corresponding mixing network. 
Finally, ELU is employed  in the mixing networks to mitigate the problem of gradient vanishing and to ensure that negative inputs remain negative (not being zeroed out as in ReLU). Indeed, ELU is convex and the two mixing networks $\cM_{\psi_R}$ and $\cM_{\psi_V}$  follow the network structure stated in Corollary~\ref{prop:convex-3}, implying that the loss function $J(\bQ,\psi_R,\psi_V)$ is convex in $\bQ$ and monotonically increasing  in each local $Q_i$.  The mixing structure in \eqref{eq:M-R} and \eqref{eq:M-V} is similar to those employed in QMIX~\citep{rashid2020monotonic}.

\subsubsection{MIFQ: Multi-agent Inverse Factorized Q-Learning Algorithm}
This section presents a practical implementation of our algorithm. Similar to \cite{garg2021iq}, we  use a $\chi^2$-regularizer $\phi(x) = x + \frac{1}{2}x^2$ for the first terms of  the loss function in  \eqref{eq:MICQ-loss-func}. 
This convex regularizer is useful to ensure that  that the term $\phi(R^{tot}(S,A))$  is lower-bounded even when the reward values go to $-\infty$, which is crucial to keep  the learning process stable.
In addition, similar to~\cite{garg2021iq}, instead of directly estimating $\bbE_{S_0}{[V^{tot}(S_0)]}$, we utilize the following equation to approximate $\bbE_{S_0}{[V^{tot}(S_0)]}$ which can stabilize training: 
\[
(1-\gamma)\bbE[V(S)] = \bbE_{(S,A)\sim \rho}\big[V(S) - \gamma \bbE_{S'\sim P(\cdot|S,A)}[V(S')]\big]
\]
for any value function $V(\cdot)$  and occupancy measure $\rho$ \citep{garg2021iq}, we can estimate $\bbE_{S_0}{[V^{tot}(S_0)]}$ by sampling $(S,A)$ from replay buffer and estimate   $\bbE_{(S,A,S')\sim \text{replay}} [V^{tot}(S)-\gamma V^{tot}(S')]$ instead. In summary, we will employ the following practical loss function:
\begin{align}
\min_{\theta,\omega_R,\omega_V} \Big\{ &J(\theta,\omega_R,\omega_V) = \sum\nolimits_{(S,A)\in \cD^E}\phi\big(R^{tot}(S,A)\big)\nonumber \\
&\quad+ \bbE_{(S,A,S')\in \cD^{replay}}\big[V^{tot}(S) -  \gamma V^{tot}(S')\big] \Big\}  \label{eq:MICQ-loss-func-practical}
\end{align}
where $\theta = (\theta_1, ..., \theta_m)$ are the training parameters of the $Q_i$ networks, and $\omega_R$ and $\omega_V$ are the parameters of the hyper-networks. 
The main steps of our MIFQ algorithm are shown in Algorithm \ref{algo:mifq}.

\subsubsection{Continuous Action Space}
The formulations and algorithm presented  above are unsuitable for continuous action spaces because the values of $V^Q_i$ in \eqref{prob:IQ-local} may not be explicitly computed w.r.t an infinite number of actions. Therefore,  we discuss in the following, an algorithm to handle this continuous-action situation. Essentially, we  consider the following maximin problem,\footnote{The weights of mixing networks are omitted for notational simplicity.} adapted from the actor-critic implementation of the single-agent IQ-Learn \cite{garg2021iq}:
\begin{align*}
\max_{\Pi}\min_{\bQ} \Big\{ &J(\bQ,\Pi) = \!\!\!\!\!\sum_{(S,A)\in \cD^E}\! \!\!\!\!\!R^{tot,\Pi}(S,A) + (1-\gamma)\bbE_{S^0}\big[V^{tot,\Pi}(S_0)\big] \Big\}  
\end{align*}
This objective depends on an alternative estimation of the local state-value functions that is suitable for the continuous action space.
\begin{align*}
    & V^{\pi_i}_i(o_i) = \bbE_{a_i\sim \pi_i(o_i\mid a_i)} [Q_i(o_i,a_i) - \log \pi_i (a_i\mid o_i)]
\end{align*}
Given this alternative estimation, we can compute the local reward functions, and joint reward and state values accordingly:
\begin{align*}
    \bR^{Q,\Pi} (S,A) &=  \bQ(S,A)-\gamma\bbE_{S'\sim P(\cdot\mid S,A)}\big[\bV^{\Pi}(S)\big] \\
    R^{tot,\Pi}(S,A) &= \cM_{\psi_R}(-\bR^{Q,\Pi}(S,A))\\
    V^{tot,\Pi}(S) &= \cM_{\psi_V}(\bV^{\Pi}(S))
\end{align*}
In other words, since the value function $V^Q$  and the agent policies $\pi^Q_i = \text{softmax}(Q_i)$ cannot be explicitly computed as functions of $Q_i$, we can  learn  an additional policy function $\Pi = [\pi_1,\ldots,\pi_m]$  and optimize it towards $\pi^Q_i$. This can be done by optimizing $J(\bQ,\Pi)$ for any fixed $\Pi$, and for any fixed $(\bQ, \psi_R,\psi_V)$,  a soft  actor-critic update can be done by solving the following optimization problem: 
\begin{equation}\label{eq:update V-pi}
\max\nolimits_{\pi_i} \Big\{ V^{\pi}_i (o_i) = \bbE_{a_i\sim \pi_i(o_i\mid a_i)} \big[Q_i(o_i,a_i) - \log \pi_i (a_i\mid o_i)\big] \Big\}    
\end{equation}
which will bring $\pi_i$ closer to $\text{softmax}(Q_i)$ \citep{garg2021iq}. Moreover,  the update in \eqref{eq:update V-pi} will bring $V^{\pi}_i$ to higher values, thus  making  every element of $\bR^{Q,\Pi}$  and $\bV^{\Pi}$ larger. Moreover, the \textit{monotonicity} of the mixing networks $\cM_{\psi_R}$ and $\cM_{\psi_V}$ then implies that such updates will also  drive the values of $R^{tot,\Pi}(S,A)$ and $V^{tot,\Pi}(S)$ upward, thereby moving towards  the goal of maximizing $J(\bQ,\Pi)$ over $\Pi$.


\section{Experiments}

\begin{figure*}[t!]
\centering
\caption{Convergence curves}
\label{fig:results}
\vspace{0.1cm}
\showlegend[0.7]{16}{4.5}{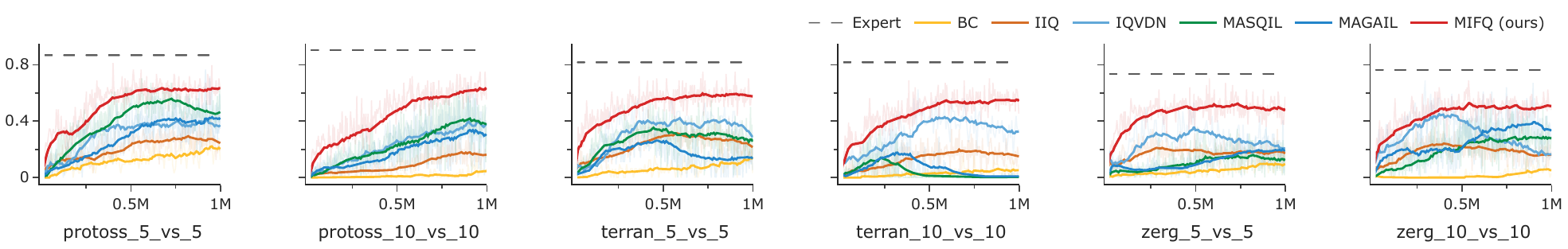} \vspace{0.15cm} \\
\showinstance[0.16]{0}{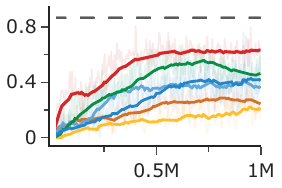}{protoss_5_vs_5}
\showinstance[0.14]{18}{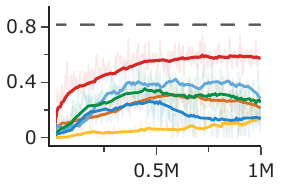}{terran_5_vs_5}
\showinstance[0.14]{18}{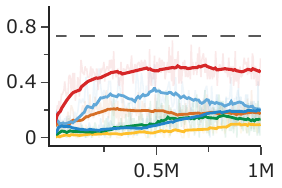}{zerg_5_vs_5}
\showinstance[0.16]{0}{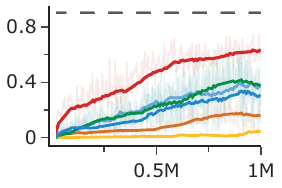}{protoss_10_vs_10}
\showinstance[0.14]{18}{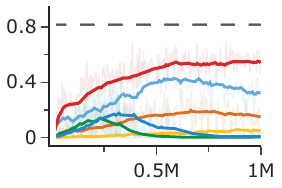}{terran_10_vs_10}
\showinstance[0.14]{18}{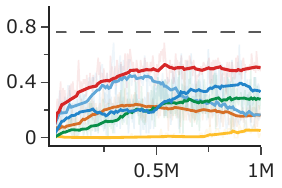}{zerg_10_vs_10} \vspace{0.2cm} \\
\showinstance[0.16]{0}{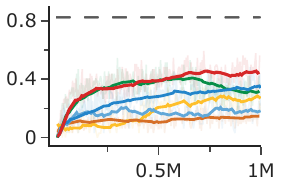}{miner_easy}
\showinstance[0.14]{18}{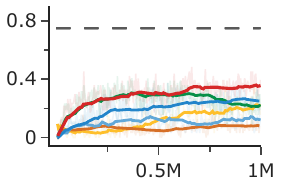}{miner_medium}
\showinstance[0.14]{18}{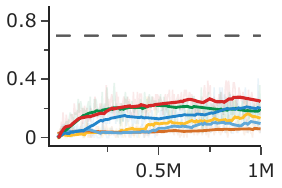}{miner_hard}
\showinstance[0.15]{0}{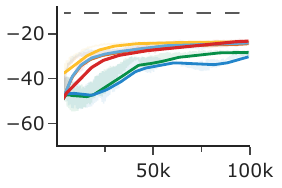}{simple_spread}
\showinstance[0.15]{0}{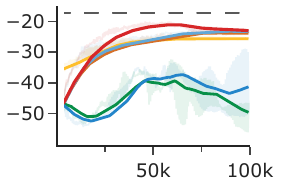}{simple_reference}
\showinstance[0.15]{0}{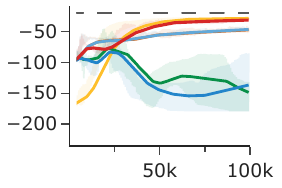}{speaker_listener}
\end{figure*}

\begin{table*}[t!]
\caption{Winrate and reward comparisons}
\label{tab:results}
\begin{tabular}{ccccccccccccc}
\toprule
\multirow{2}{*}{\text{Methods}} & \multicolumn{2}{c}{\text{Protoss}}       & \multicolumn{2}{c}{\text{Terran}}        & \multicolumn{2}{c}{\text{Zerg}}          & \multicolumn{3}{c}{\text{Miner}}                                     & \multicolumn{3}{c}{\text{MPE}}                                               \\ 
\cmidrule(l{0.4em}r{0.4em}){2-3}
\cmidrule(l{0.4em}r{0.4em}){4-5}
\cmidrule(l{0.4em}r{0.4em}){6-7}
\cmidrule(l{0.4em}r{0.4em}){8-10}
\cmidrule(l{0.4em}r{0.4em}){11-13}
                        & \multicolumn{1}{c}{\textit{5vs5}} & \textit{10vs10} & \multicolumn{1}{c}{\textit{5vs5}} & \textit{10vs10} & \multicolumn{1}{c}{\textit{5vs5}} & \textit{10vs10} & \multicolumn{1}{c}{\textit{easy}} & \multicolumn{1}{c}{\textit{medium}} & \textit{hard} & \multicolumn{1}{c}{\textit{reference}} & \multicolumn{1}{c}{\textit{spread}} & \textit{speaker} \\ \midrule
Expert                  & \multicolumn{1}{c}{87\%} & 90\%   & \multicolumn{1}{c}{82\%} & 82\%   & \multicolumn{1}{c}{74\%} & 76\%   & \multicolumn{1}{c}{82\%} & \multicolumn{1}{c}{75\%}   & 70\% & \multicolumn{1}{c}{-17.2}     & \multicolumn{1}{c}{-10.7}  & -19.7   \\
BC                      & \multicolumn{1}{c}{20\%} & 5\%    & \multicolumn{1}{c}{13\%} & 5\%    & \multicolumn{1}{c}{10\%} & 5\%    & \multicolumn{1}{c}{27\%} & \multicolumn{1}{c}{22\%}   & 13\% & \multicolumn{1}{c}{-25.6}     & \multicolumn{1}{c}{-23.6}  & \red{-28.6}   \\
IIQ                     & \multicolumn{1}{c}{25\%} & 16\%   & \multicolumn{1}{c}{22\%} & 15\%   & \multicolumn{1}{c}{17\%} & 16\%   & \multicolumn{1}{c}{15\%} & \multicolumn{1}{c}{8\%}    & 6\%  & \multicolumn{1}{c}{-23.8}     & \multicolumn{1}{c}{-24.4}  & -47.3   \\
IQVDN                   & \multicolumn{1}{c}{37\%} & 38\%   & \multicolumn{1}{c}{29\%} & 33\%   & \multicolumn{1}{c}{21\%} & 16\%   & \multicolumn{1}{c}{18\%} & \multicolumn{1}{c}{12\%}   & 10\% & \multicolumn{1}{c}{-23.2}     & \multicolumn{1}{c}{-24.1}  & -46.6   \\
MASQIL                  & \multicolumn{1}{c}{46\%} & 38\%   & \multicolumn{1}{c}{25\%} & 0\%    & \multicolumn{1}{c}{13\%} & 28\%   & \multicolumn{1}{c}{32\%} & \multicolumn{1}{c}{22\%}   & 18\% & \multicolumn{1}{c}{-49.6}     & \multicolumn{1}{c}{-28.4}  & -148.5  \\
MAGAIL                  & \multicolumn{1}{c}{41\%} & 30\%   & \multicolumn{1}{c}{13\%} & 1\%    & \multicolumn{1}{c}{19\%} & 33\%   & \multicolumn{1}{c}{34\%} & \multicolumn{1}{c}{25\%}   & 20\% & \multicolumn{1}{c}{-41.3}     & \multicolumn{1}{c}{-30.3}  & -136.5  \\
MIFQ (ours)             & \multicolumn{1}{c}{\red{63\%}} & \red{62\%}   & \multicolumn{1}{c}{\red{58\%}} & \red{54\%}   & \multicolumn{1}{c}{\red{49\%}} & \red{51\%}   & \multicolumn{1}{c}{\red{45\%}} & \multicolumn{1}{c}{\red{36\%}}   & \red{25\%} & \multicolumn{1}{c}{\red{-23.0}}     & \multicolumn{1}{c}{\red{-23.3}}  & -31.3   \\ \bottomrule
\end{tabular}
\end{table*}

\begin{figure*}[t!]
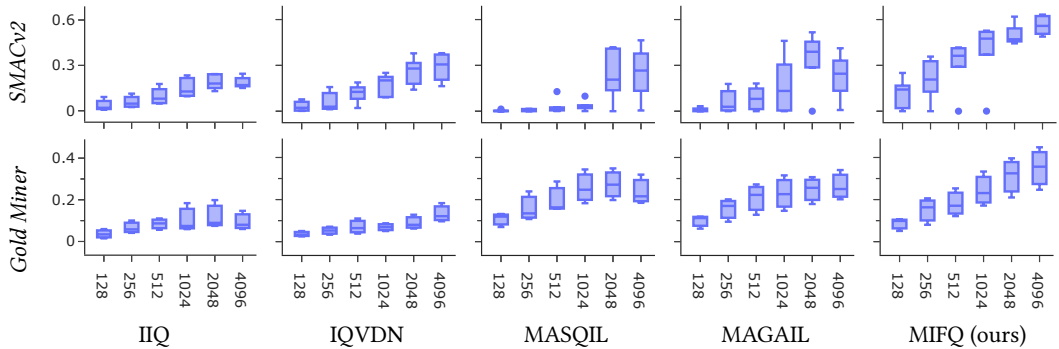

\centering
\caption{Comparison with different numbers of demonstrations. X-axis: winning rate. Y-axis: number of expert demonstrations.}
\label{fig:boxes}
\vspace{0.2cm}
\begin{tabular}{@{}p{12pt}@{}c@{}@{}c@{}@{}c@{}@{}c@{}@{}c@{}} 
\begin{tabular}[c]{@{}c@{}} \rotatebox{90}{\textit{SMACv2}} \end{tabular} &
\showinstancex[0.16]{0}{smac_iiq_box}{30} &
\showinstancex[0.14]{18}{smac_iqvdn_box}{30} &
\showinstancex[0.14]{18}{smac_sqil_box}{30} &
\showinstancex[0.14]{18}{smac_gail_box}{30} &
\showinstancex[0.14]{18}{smac_mifq_box}{30}
\\ 
\begin{tabular}[c]{@{}c@{}} \rotatebox{90}{\textit{Gold Miner}} \end{tabular} &
\showinstance[0.16]{0}{miner_iiq_box}{} &
\showinstance[0.14]{18}{miner_iqvdn_box}{} &
\showinstance[0.14]{18}{miner_sqil_box}{} &
\showinstance[0.14]{18}{miner_gail_box}{} &
\showinstance[0.14]{18}{miner_mifq_box}{}
\vspace{0.1cm} \\
& IIQ & IQVDN & MASQIL & MAGAIL & MIFQ (ours)
\end{tabular}
\end{figure*}

\subsection{Environments}
We tested our algorithm on the following three environments.

\subsubsection{SMACv2~\cite{ellis2022smacv2}} 
SMAC is a well-known multi-agent environment based on the popular real-time strategy  game named StarCraft II. In the game, 
players' objective is to collect resources, erect structures, and assemble armies of units in order to defeat their enemies. 

 In SMAC, every unit is  an autonomous agent whose actions  rely solely on local observations constrained to a limited field of view centered around that specific unit. These groups of agents need to undergo training to effectively navigate complex combat scenarios, where they face off against an opposing army that is under the centralized command of the game's pre-programmed AI.

 SMAC has become a convenient environment for evaluating the effectiveness of MARL algorithms, providing challenging multi-agent settings with characteristics reflecting real-world scenarios, including partial observability, large numbers of agents, long-term planning, complex cooperative and competitive behaviors, etc.

We employ SMACv2~\cite{ellis2022smacv2},  an enhanced version of SMAC, that introduces a more formidable environment for evaluating cooperative MARL algorithms. In SMACv2, scenarios are procedurally generated, compelling agents to adapt to previously un-encountered situations during the evaluation phase. Comparing to SMACv1~\cite{samvelyan2019starcraft}, SMACv2 allows for randomized team compositions, diverse starting positions, and a heightened focus on promoting diversity.
This benchmark comprises 6 sub-tasks, featuring a  number of agents ranging from 5 to 10. 
These agents have the flexibility to engage with opponents of differing difficulty levels. 
Within the environment, there are allies playing against enemy players. Enemies' policies are fixed and controlled by a simulator. We collect trajectories from well-trained ally agents and build IL to mimic them. The resulting imitating agents are then evaluated by letting them play against the simulator's enemies.

\subsubsection{Gold Miner~\cite{Miner}} 
This serves as another competitive multi-agent game, originating from a MARL competition.
In this game, multiple miners navigate through a 2D terrain with obstacles and deposits of gold. Players earn points based on the amount of gold they successfully collect. There are two teams, ally and enemy, playing against each other. 
  A victory  is achieved when one team mined a larger average amount of gold than the other team.
Winning this game is  challenging since the agents must adapt to competing against exceptionally well-developed heuristic-based adversaries.

We consider three sub-tasks, each involves two ally agents against two enemy agents, sorted according to three difficulty levels: (i) \textbf{Easy} (\textit{easy\_2\_vs\_2}): The enemies employ a simple shortest-path strategy to reach the gold deposits; (ii) \textbf{Medium} (\textit{medium\_2\_vs\_2}): One enemy agent follows a greedy approach, while the other emulates the algorithm employed by the second-ranking team in the competition; and (iii) \textbf{Hard} (\textit{hard\_2\_vs\_2}): This the most challenging scenario, the enemy team consists of the first- and second-ranking teams from the competition.
Finally, we collect demonstrations  from well-trained allied agents playing against enemies of fixed policies controlled by a simulator. The resulting imitators are then evaluated by letting them play against the fixed enemies.


\subsubsection{Multi Particle Environments (MPE)~\citep{lowe2017multi}}
MPE contains multiple communication-oriented deterministic multi-agent environments. We use three cooperative scenarios available in MPE for evaluating, including: (i) \textbf{Simple\_spread:} three agents learn to avoid collisions while covering all of the landmarks; (ii) \textbf{Simple\_reference:} two agents learn to get closer to the target landmarks. Each target landmark is known only by the other agents, so all agents have to communicate to each others; and (iii) \textbf{Simple\_speaker\_listener:} similar to simple\_reference, but one agent (speaker) can speak, cannot move, and one agent (listener) can move, cannot speak.

\subsection{Expert Demonstrations}

For each task, we trained an expert policy by MAPPO algorithm with large-scale hyperparameters (multi layers, higher dimensions, longer training steps, more workers, using recurrent neural network, etc.).
In term of expert buffer collection, we test each method with different numbers of expert trajectories: up to $128$ trajectories for MPEs and up to $4096$ trajectories for Miner and SMAC-v2. Note that MPEs are not dynamic environments, so we do not need a large number of expert demonstrations for evaluation.
For each collected trajectory, we used a different random seed.
After that, for a fair comparison, each method uses the same saved expert demonstrations for the training.

\subsection{Baselines}
We compare our MIFQ algorithm against other  multi-agent IL algorithms, which  either originate from the multi-agent IL literature, or be adapted from SOTA single-agent IL  algorithms.

\textbf{Behavior Cloning (\textbf{BC}).} In a multi-agent setting,  we can model each agent policy $\pi_i$ by a policy network and solving the maximum likelihood problem: $$\max_{\pi_i, i\in \cN} \sum_{i\in \cN}\sum_{(o_i,a_i)\in \cD^E} \log \pi_i(a_i|o_i).$$
  
\textbf{Independent IQ-Learn (\textbf{IIQ}).} This is  a straightforward adaption of IQ-Learn for a multi-agent setting, described in Section  \ref{sec:IndependentIQ}. 

\textbf{IQ-Learn with  Value Decomposition Network (IQVDN)}. This is similar to our MIFQ algorithm, but instead of using mixing  and hyper networks to aggregate agent Q functions, we employ the Value Decomposition (VDN)  approach \cite{sunehag2017value}.

\textbf{MASQIL.} This is our multi-agent adaption of SQIL \citep{reddy2019sqil}, a powerful framework to do imitation learning via RL. The idea is simply to assign a reward of 1 to any $(S, A)\in \cD^E$, and 0 otherwise. IL then can be done by solving an RL problem with these new rewards. In our multi-agent setting, we  adapt SQIL by employing the same reward assignment and utilize QMIX to recover policy functions. For a fair comparison, we apply the same mixing network architecture as in our MIFQ algorithm.  As a result,  MASQIL  shares some similar advantages with our MIFQ, such as being non-adversarial and enabling decentralized learning through centralized learning.

\textbf{MAGAIL.} This is a multi-agent  adversarial IL algorithm, developed by \citep{song2018multi}. 
It is worth noting that alongside MAGAIL, there is another IL algorithm, MAAIRL~\citep{yu2019multi}. MAGAIL and MAAIRL share a similar adversarial structure and are competitive in performance. Therefore, it is necessary to only select one of them for comparison. 

\subsection{Comparison Results}
We first train the imitation learning with different baselines, using 128 trajectories for MPE, and 4096 trajectories for SMACv2 and Miner.
Figure \ref{fig:results} shows the evaluation scores of each methods comparing with expert scores on each tasks during training process. In comparison, we use the wining rate metric for SMACv2 \& Gold Miner, and the reward score for MPEs, for visualization. Our method MIFQ outperforms SOTA multi-agent IL methods, i.e. MAGAIL and MAAIRL, on two hard tasks: SMACv2 \& Gold Miner; and has competitive performance on MPEs. The details are shown in Table \ref{tab:results}. Especially, on task \textit{\detokenize{protoss_5_vs_5}}, our MIFQ reaches 63\% of winning rate, higher than MAGAIL (41\%) and MASQIL (46\%), but it is still significantly lower than the expert level (87\%). On other SMACv2 tasks, and even Gold Miner tasks, the performance gaps between IL methods  and  the experts are even higher. On the easiest tasks, MPEs, MIFQ has the best scores on \textit{\detokenize{simple_reference}} and \textit{\detokenize{simple_spread}}, but slightly worse than BC on \textit{\detokenize{speaker_listener}}. Note that BC has a good performance for MPEs, which is not surprising as  MPE is a deterministic environment.

To evaluate the efficiency of our method with  different numbers of expert demonstrations/trajectories, we compare our MIFQ with MASQIL, MAGAIL, IQVDN and IIQ on two dynamic tasks: \textit{SMACv2} \& \textit{Miner}. With expert trajectories ranging  from 128 to 4096, Figure \ref{fig:boxes} shows  box and whisker plots of the average winning rate of each method on each task for data summarising analysis. As shown in the figure, our MIFQ method offers higher winrates, and it converges better with smaller standard errors. More details can be found in the appendix.

\section{Conclusion}
We developed a multi-agent IL algorithm based on the inversion of soft-Q functions. By employing mixing and hyper-network architectures, our algorithm, MIFQ, is non-adversarial and enables decentralized learning through a centralized learning approach. We demonstrated that, with some commonly used two-layer mixing network structures, our IL loss function is convex within the Q-function space, making learning convenient. Extensive experiments conducted across several challenging multi-agent tasks demonstrate the superiority of our algorithm compared to existing IL approaches. 
A potential limitation of MIFQ is that, while it achieves impressive performance across various tasks, it falls short of reaching the expertise levels. Furthermore, MIFQ, along with other baseline methods, struggles when confronted with very large-scale tasks, such as some of the largest game settings in SMACv2. Future work will focus on developing more sample-efficient and scalable IL algorithms to address these issues.




\begin{acks}
This research/project is supported by the National Research Foundation Singapore and DSO National Laboratories under the AI Singapore Programme (AISG Award No: AISG2-RP-2020-017). Dr. Nguyen was supported by grant W911NF-20-1-0344 from the US Army Research Office. 
\end{acks}



\bibliographystyle{ACM-Reference-Format} 
\bibliography{sample}

\appendix

\onecolumn

\section{Missing Proofs}

 \subsection{Proof of Theorem \ref{prop:comvex-general}}
\paragraph{Theorem  \ref{prop:comvex-general}}
\textit{ Suppose $\cM_{\psi_V}(\bX)$ and $\cM_{\psi_Q}(\bX)$ are convex in $\bX$, non-decreasing in each element $X_i$, then  the objective function $J(\bQ, \psi_V,\psi_Q)$ is convex in $\bQ$ .}
\begin{proof}
      We first denote by $\bQ$ as a vector comprising all the Q  variables: $\bQ = \{Q_i(o_i,a_i),~\forall o_i,a_i\}$. To prove  the main result, we will first verify the  following lemmas:
      \begin{lemma}\label{lm1}
    For any $i\in\ \cN$, 
    \begin{align*}
        V^Q_i(o_i) &= \log\left(\sum_{a_i} \exp(Q_i(o_i,a_i)) \right) \\
        - r^Q_i(o_i,a_i) &=  -Q_i(o_i,a_i) + \gamma \bbE_{o'_i \sim P(\cdot\mid S,A)} \big[V^Q_i(o'_i)\big]
    \end{align*}
 are convex in $\bQ$. 
 \end{lemma}
\begin{lemma}\label{lm2}
    Let $\bX(\bQ) = (X_i(\bQ), \ldots, X_m(\bQ))$ be a vector of size $m$ where each element is a convex function  of $\bQ$, i.e., $X_i(\bQ)$ is convex  in $\bQ$ for any $i\in \cN$,   $\cM_{\psi_R}(\bX(\bQ))$  and $\cM_{\psi_V}(\bX(\bQ))$  are convex in $\bQ$. 
\end{lemma}

Lemma \ref{lm1} can be easily verified, as $V^Q_i(o_i)$ has the log-sum-exp form, so it is convex  in $\bQ$ \citep{boyd2004convex}. It follows that the term $\bbE_{o'_i \sim P(\cdot\mid S,A)} \big[V^Q_i(o'_i)\big]$ is also convex in $\bQ$, implying  the convexity of $-r^Q_i(o_i,a_i)$.

 For Lemma \ref{lm2}, we will make used general properties of convex function to verify the claim, i.e., we will prove that, for any two vector $\bQ$ and $\bQ'$, and any scalar $\alpha \in (0,1)$, the following inequality always holds 
 \begin{equation}\label{eq:proof1-eq0}
 \begin{aligned}
\alpha \cM_{\psi_R}(\bX(\bQ)) + (1-\alpha) \cM_{\psi_R}(\bX(\bQ')) &\geq \cM_{\psi_R}(\bX(\alpha \bQ+ (1-\alpha)\bQ'))\\
\alpha \cM_{\psi_V}(\bX(\bQ)) + (1-\alpha) \cM_{\psi_V}(\bX(\bQ')) &\geq \cM_{\psi_V}(\bX(\alpha \bQ+ (1-\alpha)\bQ'))
 \end{aligned}
 \end{equation}
 We will validate the inequality for $\cM_{\psi_R}$; the same applies to $\cM_{}\psi_V$. According to the  convexity of $\cM_{\psi_R}(\bX)$ in $\bX$, we can write
 \begin{align}
\alpha \cM_{\psi_R}(\bX(\bQ)) + (1-\alpha) \cM_{\psi_R}(\bX(\bQ)) &\geq \cM_{\psi_R}(\alpha \bX(\bQ) + (1-\alpha)\bX(\bQ')) \\
&= \cM_{\psi_R}\begin{pmatrix}
        \alpha X_1(\bQ) +  (1-\alpha) X_1(\bQ') \\ 
        \alpha X_2(\bQ) +  (1-\alpha) X_2(\bQ') \\
        ... \\
        \alpha X_m(\bQ) +  (1-\alpha) X_m(\bQ')
    \end{pmatrix}\label{eq:proof1-eq1}
 \end{align}
 
Moreover, since $X_i(\bQ)$ is convex in $\bQ$ for all $i\in \cN$, we  have $\alpha X_i(\bQ) +  (1-\alpha) X_i(\bQ') \geq X_i(\alpha \bQ + (1-\alpha)\bQ')$. In addition, from the monotonicity of $\cM_{\psi_R}(\bX)$ in each element $X_i$, we have 
\begin{equation}\label{eq:proof1-eq2}
\cM_{\psi_R}\begin{pmatrix}
        \alpha X_1(\bQ) +  (1-\alpha) X_1(\bQ') \\ 
        \alpha X_2(\bQ) +  (1-\alpha) X_2(\bQ') \\
        ... \\
        \alpha X_m(\bQ) +  (1-\alpha) X_m(\bQ')
    \end{pmatrix} \geq \cM_{\psi_R}\begin{pmatrix}
        X_1(\alpha \bQ +  (1-\alpha) \bQ') \\ 
        X_2(\alpha \bQ +  (1-\alpha) \bQ') \\
        ... \\
        X_m(\alpha \bQ +  (1-\alpha) \bQ')
    \end{pmatrix} = \cM_{\psi_R}(\bX(\alpha\bQ +(1-\alpha)\bQ')) 
\end{equation}

Combine \eqref{eq:proof1-eq1} and \eqref{eq:proof1-eq2} we get 
\[
\alpha \cM_{\psi_R}(\bX(\bQ)) + (1-\alpha) \cM_{\psi_R}(\bX(\bQ)) \geq \cM_{\psi_R}(\bX(\alpha\bQ +(1-\alpha)\bQ'))
\]
which validates \eqref{eq:proof1-eq0}. We complete proving Lemma \ref{lm2}.

We now go back to the main result.  Considering $\cM_{\psi_R}(-\bR^Q(S,A))$, we can see that $-\bR^Q(S,A)$ is indeed a vector of size $m$  and each element  $r^Q_i(o_i,a_i)$ is convex in $\bQ$ (Lemma \ref{lm1}).  Thus,  Lemma \ref{lm2}  tells us that $\cM_{\psi_R}(-\bR^Q(S,A))$ is convex in $\bQ$. Similarly,  $\bV^Q(S)$  is also a vector of size $m$ where each element $V^Q_i(o_i)$ is convex in $\bQ$ (Lemma \ref{lm1}), thus  $\cM_{\psi_V}(\bV^Q(S))$ is  also convex in $\bQ$ .  Now, recall that the objective function of our IL is 
\[
J(\bQ,\psi_R,\psi_V) = \!\!\!\!\!\!\sum_{(S,A)\in \cD^E}\!\! \!\!\!\!\!\cM_{\psi_R}(-\bR^Q(S,A)) + (1-\gamma)\bbE_{S^0}\big[\cM_{\psi_V}(\bV^Q(S_0))\big] 
\]
which should be also convex in $\bQ$. We compete the proof.
 
\end{proof}

\subsection{Proof of Theorem \ref{prop:convex-2}}
 \paragraph{Theorem \ref{prop:convex-2}}  \textit{  Any feed-forward mixing networks $\cM_{\psi_R}(\bX)$ and  $\cM_{\psi_V}(\bX)$   constructed with  non-negative weights and nonlinear non-decreasing convex activation functions (such as \textit{ReLU } or \textit{ELU or \textit{Maxout}}) are convex  and non-decreasing in $\bX$.  }
\begin{proof}
    Any $N$-layer feed-forward network with input $\bX$ can be defined recursively as
    \begin{align}
           f^0(\bX)& = \bX \\
        f^n(\bX) &= \sigma^n\Big(f^{n-1}(\bX)\Big) \times W_n + b_n,~ n = 1,\ldots,N
    \end{align}
  where $\sigma^n$ is a set of activation functions applied to each element of vector $f^{n-1}(\bX)$, and $W_n$ and $b_n$ are the weights  and biases,  respectively, at layer $n$.  Therefore, we will prove the result by induction, i.e., $f^n(\bX)$ is convex and non-decreasing in $\bX$  for $n = 0,\ldots$. Here we note that $f^(\bX)$ is a vector, so when we say ``\textit{$f^n(\bX)$ is convex and non-decreasing in $\bX$}'', it implies each element of $f^n(\bX)$ is convex and non-decreasing in $\bX$.
  
  We first see that claim indeed holds for $n = 0$ . Now let us assume that,  $f^{n-1}(\bX)$ is convex  and non-decreasing in $\bX$, we will prove that $f^n(\bX)$ is also convex and non-decreasing in $\bX$. The non-decreasing property  can be easily verified as we can see, given two vector $\bX$ and $\bX'$ such that $\bX\geq\bX'$  (element-wise comparison),  then we have the following chain of inequalities
  \begin{align*}
      f^{n-1}(\bX) &\stackrel{(a)}{\geq} f^{n-1}(\bX') \\
      \sigma^n(f^{n-1}(\bX)) &\stackrel{(b)}{\geq} \sigma^n(f^{n-1}(\bX'))\\
      \sigma^n(f^{n-1}(\bX)) \times W_n + b_n &\stackrel{(c)}{\geq} \sigma^n(f^{n-1}(\bX')) \times W_n + b_n
  \end{align*}

where $(a)$ is due to the  induction assumption that $f^{n-1}(\bX)$ is non-decreasing in $\bX$, $(b)$ is because $\sigma^n$ are also non-decreasing, and $(c)$ is because the weights $W_n$ is non-negative.

To verify the convexity of $f^n(\bX)$,we will show that for any $\bX,\bX'$,  and any scalar $\alpha \in (0,1)$,  the following holds 
\begin{equation}
   \alpha f^n(\bX) + (1-\alpha) f^n(\bX) \geq f^n(\alpha \bX + (1-\alpha)\bX')    
\end{equation}
To this end, we write 
\begin{align*}
     \alpha f^n(\bX) + (1-\alpha) f^n(\bX')& = \Big(\alpha\sigma^n(f^{n-1}(\bX))+ (1-\alpha)\sigma^n(f^{n-1}(\bX'))\Big) \times W_n + b_n\\
     &\stackrel{(d)}{\geq} \Big(\sigma^n\Big(\alpha f^{n-1}(\bX)+ (1-\alpha)f^{n-1}(\bX')\Big) \times W_n + b_n \\
     &\stackrel{(e)}{\geq} \Big(\sigma^n\Big( f^{n-1}(\alpha \bX+ (1-\alpha)\bX')\Big) \times W_n + b_n \\
     &= f^n(\alpha \bX+ (1-\alpha)\bX')
\end{align*}
where $(d)$ is due to the assumption that activation functions $\sigma^n$ are convex and $W_n\geq 0$, and $(e)$ is because  $\alpha f^{n-1}(\bX)+ (1-\alpha)f^{n-1}(\bX')\geq f^{n-1}(\alpha \bX+ (1-\alpha)\bX')$ (because $f^{n-1}(\bX)$ is convex in $\bX$, by the induction assumption) and the activation functions $\sigma^n$ is non-decreasing and $W_n\geq 0$.  So, we have
\[
  \alpha f^n(\bX) + (1-\alpha) f^n(\bX') \geq f^n(\alpha \bX+ (1-\alpha)\bX')
\]
implying that $f^n(\bX)$ is convex in $\bX$. We then complete the induction proof and conclude that $f^n(\bX)$  is convex and non-decreasing in $\bX$ for any $n = 0,...,N$.

\end{proof}

\subsection{Proof  of Proposition \ref{prop:linear-mixing}}

\textbf{Proposition \ref{prop:linear-mixing}:}
\textit{If $\cM_{\psi_R}(\bX)$  and $\cM_{\psi_V}(\bX)$ are weighted sums of $X_i$ with non-negative weighting parameters, i.e., $\cM_{\psi_R}(\bX) = \cM_{\psi_V}(\bX) = \sum_{i}\alpha_i X_i$, $\alpha_i\geq 0$ ($\forall i\in \cN$),  then  $J(\bQ,\psi_R,\psi_V)$ is convex in $\bQ$. Moreover, minimizing $J(\bQ,\psi_R,\psi_V)$ over the $\bQ$ space is equivalent to minimizing  each local loss function $J_i(Q_i)$. That is,
\begin{align*}
    \text{argmin}_\bQ \{J(\bQ,\psi_R,\psi_V)\} = \big[\text{argmin}_{Q_1} J_1(Q_1),\cdots, \text{argmin}_{Q_m} J_m(Q_m)\big]
\end{align*}}

\begin{proof}
    Under the  mixing networks defined in Proposition \ref{prop:linear-mixing}, we write the objective function as
    \begin{align}
        J(\bQ,\psi_R,\psi_V) &= 
        \sum_{(S,A)\in \cD^E}\!\! 
        \cM_{\psi_R}(-\bR^Q(S,A)) + (1-\gamma)\bbE_{S^0}\big[\cM_{\psi_V}(\bV^Q(S_0))\big] \\
        &=
        \sum_{(S,A)\in \cD^E}
        \sum_{i\in \cN}-\alpha_i r^Q_i(o_i,a_i) + (1-\gamma)\bbE_{S^0}\big[\sum_{i\in \cN}\alpha_i(V^Q_i(o^0_{i}))\big] \\
        &= 
        \sum_{i\in \cN}
        \alpha_i J_i(Q_i)
    \end{align}
    Since $\alpha_i\geq 0$ and $Q_i$ are independent from each other, to minimize $J(\bQ,\psi_R,\psi_V)$, each component  $J_i(Q_i)$ needs to be minimized, which directly leads to the desired result. 

    It is important to note that the above result only holds if $Q_i$ are independent. It is not the case if $Q_i$, for some $i$,  share a common network structure. This is also the case of the IQVDN considered in the main paper, i.e., the global reward and value function $R^{tot}$ and $V^{tot}$ are sums of the corresponding local functions, but $Q_i$ share the same neural network structure. 
\end{proof}

\section{Additional Details}

\subsection{Network Architecture }

\begin{figure*}[ht]
    \centering
    \caption{An overview of our network architecture; The mixing and hyper-networks of $R^{tot}$ are similar to those of $V^{tot}$.}
    \vspace{0.4cm}
    \label{fig:model}
    \includegraphics[width=0.9\textwidth]{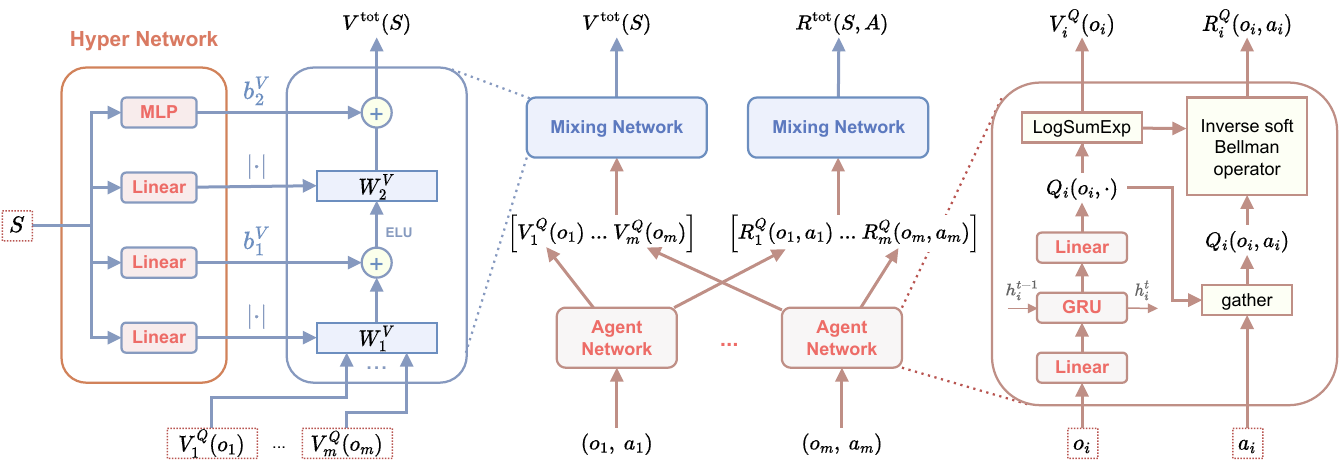}
    \label{fig.diagram}
\end{figure*}

Figure \ref{fig:model}  presents an overview of our neural network architecture, including illustrations for the $Q_i$, mixing  and hyper networks.

\subsection{Experimental Settings}
Each environment has different a observation space, state dimension, and action space. Therefore, we use different hyper-parameters on each task to try to make all algorithms work stably. Moreover, due to limitations in computing resources, especially random access memory, we reduce the buffer size to 5000 on two hardest tasks, Miner and SMACv2, to be more efficient in running time with parallel workers. More details are available in Table \ref{tab:hyperparams}. We use four High-Performance Computing (HPC) clusters for training and evaluating all tasks. Specifically, each HPC cluster has a workload with an NVIDIA L40 GPU 48 GB GDDR6, 32 Intel-CPU cores, and 100GB RAM.
In terms of model architecture, Figure \ref{fig:model} shows our proposed model structure with mixing networks based on QMIX algorithm.

\begin{table*}[ht]
\caption{Hyper-parameters.}
\label{tab:hyperparams}
\begin{tabular}{c|ccc}
\toprule
\textit{Arguments}                   & \multicolumn{1}{c|}{\textit{MPEs}}    & \multicolumn{1}{c|}{\textit{Miner}} & \textit{SMACv2} \\ \midrule
Max training steps      & \multicolumn{1}{c|}{100000} & \multicolumn{2}{c}{1000000}        \\ 
Evaluate times          & \multicolumn{3}{c}{32}                                           \\ 
Buffer size             & \multicolumn{1}{c|}{100000} & \multicolumn{2}{c}{5000}           \\ 
Learning rate           & \multicolumn{1}{c|}{2e-5}   & \multicolumn{2}{c}{5e-4}           \\ 
Batch size              & \multicolumn{3}{c}{128}                                          \\ 
Hidden dim              & \multicolumn{3}{c}{128}                                          \\ 
Gamma                   & \multicolumn{3}{c}{0.99}                                         \\ 
Target update frequency & \multicolumn{3}{c}{4}                                            \\ 
Number of random seeds         & \multicolumn{3}{c}{4}                                            \\ \bottomrule
\end{tabular}
\end{table*}

\subsection{Experimental Details}

In this section, we present in detail experimental results for SMACv2, Miner, and MPE tasks, with varying  numbers of expert demonstrations.

\showsmactable{128}{
IIQ & 9.1\% & \multicolumn{1}{c}{1.2\%} & 6.5\% & \multicolumn{1}{c}{1.2\%} & 2.8\% & \multicolumn{1}{c}{0.8\%} & 3.6\% \\ 
IQVDN & 6.1\% & \multicolumn{1}{c}{0.2\%} & 7.8\% & \multicolumn{1}{c}{0.5\%} & 3.9\% & \multicolumn{1}{c}{0.4\%} & 3.1\% \\ 
MASQIL & 0.3\% & \multicolumn{1}{c}{0.0\%} & 0.3\% & \multicolumn{1}{c}{0.0\%} & 1.4\% & \multicolumn{1}{c}{0.0\%} & 0.3\% \\ 
MAGAIL & 0.6\% & \multicolumn{1}{c}{0.0\%} & 0.0\% & \multicolumn{1}{c}{0.0\%} & 3.3\% & \multicolumn{1}{c}{1.7\%} & 0.9\% \\ 
MIFQ (ours) & 15.9\% & \multicolumn{1}{c}{16.8\%} & 25.1\% & \multicolumn{1}{c}{0.0\%} & 2.0\% & \multicolumn{1}{c}{12.5\%} & 12.0\% \\ 
Expert & 86.7\% & \multicolumn{1}{c}{90.3\%} & 81.7\% & \multicolumn{1}{c}{81.7\%} & 73.5\% & \multicolumn{1}{c}{76.3\%} & 81.7\% \\ 
}

\showsmactable{256}{
IIQ & 11.4\% & \multicolumn{1}{c}{2.7\%} & 6.7\% & \multicolumn{1}{c}{2.4\%} & 9.1\% & \multicolumn{1}{c}{2.7\%} & 5.8\% \\ 
IQVDN & 11.9\% & \multicolumn{1}{c}{1.6\%} & 15.8\% & \multicolumn{1}{c}{1.3\%} & 3.9\% & \multicolumn{1}{c}{2.0\%} & 6.1\% \\ 
MASQIL & 1.0\% & \multicolumn{1}{c}{0.1\%} & 0.0\% & \multicolumn{1}{c}{0.0\%} & 1.6\% & \multicolumn{1}{c}{1.4\%} & 0.7\% \\ 
MAGAIL & 13.1\% & \multicolumn{1}{c}{17.9\%} & 0.5\% & \multicolumn{1}{c}{0.0\%} & 4.3\% & \multicolumn{1}{c}{1.5\%} & 6.2\% \\ 
MIFQ (ours) & 35.8\% & \multicolumn{1}{c}{32.7\%} & 26.9\% & \multicolumn{1}{c}{0.0\%} & 12.8\% & \multicolumn{1}{c}{14.6\%} & 20.5\% \\ 
Expert & 86.7\% & \multicolumn{1}{c}{90.3\%} & 81.7\% & \multicolumn{1}{c}{81.7\%} & 73.5\% & \multicolumn{1}{c}{76.3\%} & 81.7\% \\ 
}

\showsmactable{512}{
IIQ & 17.7\% & \multicolumn{1}{c}{7.0\%} & 14.3\% & \multicolumn{1}{c}{5.1\%} & 9.0\% & \multicolumn{1}{c}{4.9\%} & 9.7\% \\ 
IQVDN & 18.8\% & \multicolumn{1}{c}{8.1\%} & 15.6\% & \multicolumn{1}{c}{15.1\%} & 10.2\% & \multicolumn{1}{c}{2.1\%} & 11.7\% \\ 
MASQIL & 0.1\% & \multicolumn{1}{c}{0.7\%} & 2.5\% & \multicolumn{1}{c}{0.5\%} & 12.8\% & \multicolumn{1}{c}{1.8\%} & 3.1\% \\ 
MAGAIL & 1.6\% & \multicolumn{1}{c}{18.2\%} & 14.6\% & \multicolumn{1}{c}{0.0\%} & 14.8\% & \multicolumn{1}{c}{1.4\%} & 8.4\% \\ 
MIFQ (ours) & 41.7\% & \multicolumn{1}{c}{37.8\%} & 41.2\% & \multicolumn{1}{c}{0.0\%} & 29.1\% & \multicolumn{1}{c}{34.7\%} & 30.8\% \\ 
Expert & 86.7\% & \multicolumn{1}{c}{90.3\%} & 81.7\% & \multicolumn{1}{c}{81.7\%} & 73.5\% & \multicolumn{1}{c}{76.3\%} & 81.7\% \\ 
}

\showsmactable{1024}{
IIQ & 23.4\% & \multicolumn{1}{c}{13.5\%} & 21.6\% & \multicolumn{1}{c}{9.6\%} & 9.9\% & \multicolumn{1}{c}{12.0\%} & 15.0\% \\ 
IQVDN & 20.1\% & \multicolumn{1}{c}{25.1\%} & 22.5\% & \multicolumn{1}{c}{20.2\%} & 9.1\% & \multicolumn{1}{c}{9.3\%} & 17.7\% \\ 
MASQIL & 3.8\% & \multicolumn{1}{c}{3.9\%} & 0.1\% & \multicolumn{1}{c}{2.4\%} & 9.8\% & \multicolumn{1}{c}{2.0\%} & 3.7\% \\ 
MAGAIL & 46.1\% & \multicolumn{1}{c}{26.0\%} & 0.5\% & \multicolumn{1}{c}{0.0\%} & 30.5\% & \multicolumn{1}{c}{0.5\%} & 17.3\% \\ 
MIFQ (ours) & 50.3\% & \multicolumn{1}{c}{52.7\%} & 51.9\% & \multicolumn{1}{c}{0.1\%} & 37.0\% & \multicolumn{1}{c}{44.7\%} & 39.4\% \\ 
Expert & 86.7\% & \multicolumn{1}{c}{90.3\%} & 81.7\% & \multicolumn{1}{c}{81.7\%} & 73.5\% & \multicolumn{1}{c}{76.3\%} & 81.7\% \\ 
}

\showsmactable{2048}{
IIQ & 24.2\% & \multicolumn{1}{c}{15.0\%} & 17.8\% & \multicolumn{1}{c}{24.3\%} & 13.0\% & \multicolumn{1}{c}{18.2\%} & 18.7\% \\ 
IQVDN & 29.9\% & \multicolumn{1}{c}{25.9\%} & 31.6\% & \multicolumn{1}{c}{37.9\%} & 18.0\% & \multicolumn{1}{c}{14.1\%} & 26.2\% \\ 
MASQIL & 41.0\% & \multicolumn{1}{c}{41.9\%} & 13.7\% & \multicolumn{1}{c}{0.0\%} & 19.9\% & \multicolumn{1}{c}{21.4\%} & 23.0\% \\ 
MAGAIL & 51.7\% & \multicolumn{1}{c}{45.1\%} & 28.7\% & \multicolumn{1}{c}{0.0\%} & 32.7\% & \multicolumn{1}{c}{45.5\%} & 33.9\% \\ 
MIFQ (ours) & 61.9\% & \multicolumn{1}{c}{47.1\%} & 54.2\% & \multicolumn{1}{c}{45.7\%} & 44.4\% & \multicolumn{1}{c}{47.0\%} & 50.1\% \\ 
Expert & 86.7\% & \multicolumn{1}{c}{90.3\%} & 81.7\% & \multicolumn{1}{c}{81.7\%} & 73.5\% & \multicolumn{1}{c}{76.3\%} & 81.7\% \\ 
}

\showsmactable{4096}{
BC & 20.2\% & \multicolumn{1}{c}{4.6\%} & 13.1\% & \multicolumn{1}{c}{5.2\%} & 9.6\% & \multicolumn{1}{c}{5.1\%} & 9.6\% \\ 
IIQ & 24.5\% & \multicolumn{1}{c}{16.2\%} & 21.6\% & \multicolumn{1}{c}{15.1\%} & 16.9\% & \multicolumn{1}{c}{16.4\%} & 18.5\% \\ 
IQVDN & 37.1\% & \multicolumn{1}{c}{38.0\%} & 28.5\% & \multicolumn{1}{c}{32.7\%} & 20.6\% & \multicolumn{1}{c}{16.4\%} & 28.9\% \\ 
MASQIL & 46.4\% & \multicolumn{1}{c}{37.7\%} & 25.5\% & \multicolumn{1}{c}{0.5\%} & 13.4\% & \multicolumn{1}{c}{27.7\%} & 25.2\% \\ 
MAGAIL & 41.2\% & \multicolumn{1}{c}{29.7\%} & 13.4\% & \multicolumn{1}{c}{0.8\%} & 19.4\% & \multicolumn{1}{c}{33.1\%} & 22.9\% \\ 
MIFQ (ours) & 63.2\% & \multicolumn{1}{c}{62.3\%} & 57.6\% & \multicolumn{1}{c}{54.1\%} & 48.8\% & \multicolumn{1}{c}{50.7\%} & 56.1\% \\ 
Expert & 86.7\% & \multicolumn{1}{c}{90.3\%} & 81.7\% & \multicolumn{1}{c}{81.7\%} & 73.5\% & \multicolumn{1}{c}{76.3\%} & 81.7\% \\ 
}

\showminertable{128}{
IIQ & \multicolumn{1}{c}{6.0\%} & \multicolumn{1}{c}{3.4\%} & 1.5\% & 3.6\% \\
IQVDN & \multicolumn{1}{c}{3.2\%} & \multicolumn{1}{c}{5.1\%} & 2.5\% & 3.6\% \\
MASQIL & \multicolumn{1}{c}{13.2\%} & \multicolumn{1}{c}{11.9\%} & 7.0\% & 10.7\% \\
MAGAIL & \multicolumn{1}{c}{11.8\%} & \multicolumn{1}{c}{11.7\%} & 6.2\% & 9.9\% \\
MIFQ (ours) & \multicolumn{1}{c}{10.0\%} & \multicolumn{1}{c}{10.6\%} & 5.1\% & 8.6\% \\
Expert & \multicolumn{1}{c}{82.4\%} & \multicolumn{1}{c}{74.9\%} & 69.8\% & 75.7\% \\
}

\showminertable{256}{
IIQ & \multicolumn{1}{c}{10.1\%} & \multicolumn{1}{c}{5.8\%} & 4.1\% & 6.7\% \\
IQVDN & \multicolumn{1}{c}{7.1\%} & \multicolumn{1}{c}{5.3\%} & 3.3\% & 5.2\% \\
MASQIL & \multicolumn{1}{c}{23.9\%} & \multicolumn{1}{c}{13.5\%} & 10.7\% & 16.0\% \\
MAGAIL & \multicolumn{1}{c}{20.1\%} & \multicolumn{1}{c}{17.0\%} & 9.5\% & 15.5\% \\
MIFQ (ours) & \multicolumn{1}{c}{20.6\%} & \multicolumn{1}{c}{16.3\%} & 8.0\% & 15.0\% \\
Expert & \multicolumn{1}{c}{82.4\%} & \multicolumn{1}{c}{74.9\%} & 69.8\% & 75.7\% \\
}

\showminertable{512}{
IIQ & \multicolumn{1}{c}{11.1\%} & \multicolumn{1}{c}{8.7\%} & 5.6\% & 8.5\% \\
IQVDN & \multicolumn{1}{c}{11.0\%} & \multicolumn{1}{c}{6.4\%} & 3.8\% & 7.1\% \\
MASQIL & \multicolumn{1}{c}{28.6\%} & \multicolumn{1}{c}{15.9\%} & 15.7\% & 20.1\% \\
MAGAIL & \multicolumn{1}{c}{27.2\%} & \multicolumn{1}{c}{22.2\%} & 12.8\% & 20.7\% \\
MIFQ (ours) & \multicolumn{1}{c}{25.3\%} & \multicolumn{1}{c}{17.1\%} & 12.1\% & 18.2\% \\
Expert & \multicolumn{1}{c}{82.4\%} & \multicolumn{1}{c}{74.9\%} & 69.8\% & 75.7\% \\
}

\showminertable{1024}{
IIQ & \multicolumn{1}{c}{18.4\%} & \multicolumn{1}{c}{7.5\%} & 5.9\% & 10.6\% \\
IQVDN & \multicolumn{1}{c}{8.7\%} & \multicolumn{1}{c}{5.0\%} & 6.9\% & 6.9\% \\
MASQIL & \multicolumn{1}{c}{34.3\%} & \multicolumn{1}{c}{24.7\%} & 18.3\% & 25.8\% \\
MAGAIL & \multicolumn{1}{c}{31.5\%} & \multicolumn{1}{c}{22.6\%} & 14.7\% & 22.9\% \\
MIFQ (ours) & \multicolumn{1}{c}{33.3\%} & \multicolumn{1}{c}{23.1\%} & 17.1\% & 24.5\% \\
Expert & \multicolumn{1}{c}{82.4\%} & \multicolumn{1}{c}{74.9\%} & 69.8\% & 75.7\% \\
}

\showminertable{2048}{
IIQ & \multicolumn{1}{c}{19.9\%} & \multicolumn{1}{c}{8.9\%} & 7.5\% & 12.1\% \\
IQVDN & \multicolumn{1}{c}{12.8\%} & \multicolumn{1}{c}{8.0\%} & 6.2\% & 9.0\% \\
MASQIL & \multicolumn{1}{c}{34.8\%} & \multicolumn{1}{c}{27.1\%} & 19.8\% & 27.2\% \\
MAGAIL & \multicolumn{1}{c}{30.7\%} & \multicolumn{1}{c}{25.6\%} & 17.9\% & 24.7\% \\
MIFQ (ours) & \multicolumn{1}{c}{39.6\%} & \multicolumn{1}{c}{32.5\%} & 21.0\% & 31.0\% \\
Expert & \multicolumn{1}{c}{82.4\%} & \multicolumn{1}{c}{74.9\%} & 69.8\% & 75.7\% \\
}

\showminertable{4096}{
BC & \multicolumn{1}{c}{26.9\%} & \multicolumn{1}{c}{21.7\%} & 13.3\% & 20.6\% \\
IIQ & \multicolumn{1}{c}{14.7\%} & \multicolumn{1}{c}{8.0\%} & 6.0\% & 9.6\% \\
IQVDN & \multicolumn{1}{c}{18.3\%} & \multicolumn{1}{c}{12.1\%} & 9.7\% & 13.4\% \\
MASQIL & \multicolumn{1}{c}{31.9\%} & \multicolumn{1}{c}{21.6\%} & 18.5\% & 24.0\% \\
MAGAIL & \multicolumn{1}{c}{34.0\%} & \multicolumn{1}{c}{25.0\%} & 20.1\% & 26.4\% \\
MIFQ (ours) & \multicolumn{1}{c}{44.9\%} & \multicolumn{1}{c}{35.6\%} & 24.7\% & 35.0\% \\
Expert & \multicolumn{1}{c}{82.4\%} & \multicolumn{1}{c}{74.9\%} & 69.8\% & 75.7\% \\
}

\showmpetable{1}{
IIQ & \multicolumn{1}{c}{-27.8} & \multicolumn{1}{c}{-38.4} & -64.9 & -43.7 \\
MASQIL & \multicolumn{1}{c}{-45.2} & \multicolumn{1}{c}{-56.4} & -128.3 & -76.6 \\
MAGAIL & \multicolumn{1}{c}{-50.9} & \multicolumn{1}{c}{-43.9} & -101.6 & -65.5 \\
MIFQ (ours) & \multicolumn{1}{c}{-49.5} & \multicolumn{1}{c}{-37.5} & -78.5 & -55.2 \\
Expert & \multicolumn{1}{c}{-17.2} & \multicolumn{1}{c}{-10.7} & -19.7 & -15.8 \\
}

\showmpetable{2}{
IIQ & \multicolumn{1}{c}{-28.5} & \multicolumn{1}{c}{-34.5} & -80.6 & -47.9 \\
MASQIL & \multicolumn{1}{c}{-48.4} & \multicolumn{1}{c}{-54.2} & -114.3 & -72.3 \\
MAGAIL & \multicolumn{1}{c}{-47.3} & \multicolumn{1}{c}{-42.7} & -104.1 & -64.7 \\
MIFQ (ours) & \multicolumn{1}{c}{-45.9} & \multicolumn{1}{c}{-38.9} & -88.6 & -57.8 \\
Expert & \multicolumn{1}{c}{-17.2} & \multicolumn{1}{c}{-10.7} & -19.7 & -15.8 \\
}

\showmpetable{4}{
IIQ & \multicolumn{1}{c}{-29.0} & \multicolumn{1}{c}{-33.1} & -67.4 & -43.1 \\
MASQIL & \multicolumn{1}{c}{-47.9} & \multicolumn{1}{c}{-44.5} & -112.0 & -68.1 \\
MAGAIL & \multicolumn{1}{c}{-49.4} & \multicolumn{1}{c}{-43.3} & -90.2 & -61.0 \\
MIFQ (ours) & \multicolumn{1}{c}{-46.1} & \multicolumn{1}{c}{-37.3} & -119.6 & -67.6 \\
Expert & \multicolumn{1}{c}{-17.2} & \multicolumn{1}{c}{-10.7} & -19.7 & -15.8 \\
}

\showmpetable{8}{
IIQ & \multicolumn{1}{c}{-27.7} & \multicolumn{1}{c}{-27.7} & -62.7 & -39.4 \\
MASQIL & \multicolumn{1}{c}{-43.7} & \multicolumn{1}{c}{-49.6} & -110.5 & -68.0 \\
MAGAIL & \multicolumn{1}{c}{-41.2} & \multicolumn{1}{c}{-43.9} & -99.5 & -61.5 \\
MIFQ (ours) & \multicolumn{1}{c}{-44.1} & \multicolumn{1}{c}{-34.3} & -61.1 & -46.5 \\
Expert & \multicolumn{1}{c}{-17.2} & \multicolumn{1}{c}{-10.7} & -19.7 & -15.8 \\
}

\showmpetable{16}{
IIQ & \multicolumn{1}{c}{-27.9} & \multicolumn{1}{c}{-27.2} & -57.2 & -37.4 \\
MASQIL & \multicolumn{1}{c}{-44.2} & \multicolumn{1}{c}{-45.8} & -128.5 & -72.8 \\
MAGAIL & \multicolumn{1}{c}{-42.4} & \multicolumn{1}{c}{-40.4} & -135.1 & -72.6 \\
MIFQ (ours) & \multicolumn{1}{c}{-51.8} & \multicolumn{1}{c}{-31.5} & -42.1 & -41.8 \\
Expert & \multicolumn{1}{c}{-17.2} & \multicolumn{1}{c}{-10.7} & -19.7 & -15.8 \\
}

\showmpetable{32}{
IIQ & \multicolumn{1}{c}{-25.7} & \multicolumn{1}{c}{-27.4} & -46.3 & -33.1 \\
MASQIL & \multicolumn{1}{c}{-45.4} & \multicolumn{1}{c}{-34.7} & -105.9 & -62.0 \\
MAGAIL & \multicolumn{1}{c}{-36.3} & \multicolumn{1}{c}{-34.1} & -133.5 & -68.0 \\
MIFQ (ours) & \multicolumn{1}{c}{-31.8} & \multicolumn{1}{c}{-25.2} & -36.7 & -31.3 \\
Expert & \multicolumn{1}{c}{-17.2} & \multicolumn{1}{c}{-10.7} & -19.7 & -15.8 \\
}

\showmpetable{64}{
IIQ & \multicolumn{1}{c}{-23.9} & \multicolumn{1}{c}{-24.5} & -48.5 & -32.3 \\
MASQIL & \multicolumn{1}{c}{-37.1} & \multicolumn{1}{c}{-32.2} & -146.3 & -71.8 \\
MAGAIL & \multicolumn{1}{c}{-41.7} & \multicolumn{1}{c}{-32.1} & -117.6 & -63.8 \\
MIFQ (ours) & \multicolumn{1}{c}{-24.4} & \multicolumn{1}{c}{-23.9} & -33.5 & -27.3 \\
Expert & \multicolumn{1}{c}{-17.2} & \multicolumn{1}{c}{-10.7} & -19.7 & -15.8 \\
}

\showmpetable{128}{
BC & \multicolumn{1}{c}{-25.6} & \multicolumn{1}{c}{-23.6} & -28.6 & -25.9 \\
IIQ & \multicolumn{1}{c}{-23.8} & \multicolumn{1}{c}{-24.4} & -47.3 & -31.8 \\
IQVDN & \multicolumn{1}{c}{-23.2} & \multicolumn{1}{c}{-24.1} & -46.6 & -31.3 \\
MASQIL & \multicolumn{1}{c}{-49.6} & \multicolumn{1}{c}{-28.4} & -148.5 & -75.5 \\
MAGAIL & \multicolumn{1}{c}{-41.3} & \multicolumn{1}{c}{-30.3} & -136.5 & -69.4 \\
MIFQ (ours) & \multicolumn{1}{c}{-23.0} & \multicolumn{1}{c}{-23.3} & -31.3 & -25.8 \\
Expert & \multicolumn{1}{c}{-17.2} & \multicolumn{1}{c}{-10.7} & -19.7 & -15.8 \\
}

\end{document}